\documentclass{article}
\usepackage{ijcai25}

\usepackage{times}
\usepackage{soul}
\usepackage{url}
\usepackage[hidelinks]{hyperref}
\usepackage{xcolor}
\hypersetup{
colorlinks=true,
linkcolor=[HTML]{B2001D},
citecolor=[HTML]{27408B},
urlcolor=[HTML]{27408B}
}
\usepackage{orcidlink}
\usepackage[utf8]{inputenc}
\usepackage[small]{caption}
\usepackage{graphicx}
\usepackage{amsmath}
\usepackage{amsthm}
\usepackage{booktabs}
\usepackage{algorithm}
\usepackage{algorithmic}
\usepackage[switch]{lineno}

\usepackage{amssymb}
\usepackage{amsfonts}
\usepackage{multirow}
\usepackage{colortab}
\usepackage{colortbl}
\usepackage{arydshln}
\usepackage{subcaption}
\newcommand{\bhline}[1]{\noalign{\hrule height #1}}   
\definecolor{g}{rgb}{0.925, 0.957, 0.831} 
\definecolor{p}{rgb}{0.980,0.910,0.922}
\newcommand{\pop}{\mathcal{P}}
\newcommand{\ms}{\mathcal{M}}
\newcommand{\pc}{\mathcal{P}_{C}}

\usepackage{pgfplots}
 \pgfplotsset{compat=1.18}

\title{X-KAN: Optimizing Local Kolmogorov-Arnold Networks via\\ Evolutionary Rule-Based Machine Learning}

\iftrue
\author{
Hiroki Shiraishi$^1$\orcidlink{0000-0001-8730-1276}
\and
Hisao Ishibuchi$^2$\footnote{Corresponding authors.}\orcidlink{0000-0001-9186-6472}\And
Masaya Nakata$^{1}$\footnotemark[1]\orcidlink{0000-0003-3428-7890}\\
\affiliations
$^1$Faculty of Engineering, Yokohama National University\\
$^2$Department of Computer Science and Engineering, Southern University of Science and Technology\\
\emails
shiraishi-hiroki-yw@ynu.jp,
hisao@sustech.edu.cn,
nakata-masaya-tb@ynu.ac.jp
}
\fi

\begin{document}

\maketitle

\begin{abstract}
Function approximation is a critical task in various fields. However, existing neural network approaches struggle with locally complex or discontinuous functions due to their reliance on a single global model covering the entire problem space. We propose X-KAN, a novel method that optimizes multiple local Kolmogorov-Arnold Networks (KANs) through an evolutionary rule-based machine learning framework called XCSF. X-KAN combines KAN's high expressiveness with XCSF's adaptive partitioning capability by implementing local KAN models as rule consequents and defining local regions via rule antecedents. Our experimental results on artificial test functions and real-world datasets demonstrate that X-KAN significantly outperforms conventional methods, including XCSF, Multi-Layer Perceptron, and KAN, in terms of approximation accuracy. Notably, X-KAN effectively handles functions with locally complex or discontinuous structures that are challenging for conventional KAN, using a compact set of rules (average 7.2 $\pm$ 2.3 rules). These results validate the effectiveness of using KAN as a local model in XCSF, which evaluates the rule fitness based on both accuracy and generality. Our X-KAN implementation is available at \url{https://github.com/YNU-NakataLab/X-KAN}.

\end{abstract}

\section{Introduction}

Function approximation is a crucial task in various industrial fields, including control system design and signal processing \cite{bao2022adaptive,mirza2024decoding}. The function approximation problem addressed in this paper aims to discover the most suitable approximation function for given data points, and can be formalized as follows:
\begin{align}
    \text{Given: }& \mathcal{D}=\{(\mathbf{x}_i, y_i) \in [0,1]^n \times \mathbb{R}\mid i = 1, \ldots, M\}, \\
    \text{Find: }& \hat{f} = \arg\min_{\hat{f}} \sum_{i=1}^M \mathcal{L}\left(y_i,\hat{f}(\mathbf{x}_i)\right),
\end{align}
where $\mathcal{D}$ is a training dataset, $n$ is the input dimensionality, $M$ is the total number of data points in $\mathcal{D}$, $\hat{f}: \mathbb{R}^n \rightarrow \mathbb{R}$ is the approximation function, and $\mathcal{L}: \mathbb{R} \times \mathbb{R} \rightarrow \mathbb{R}_0^+$ is the loss function. The difficulty of this problem becomes particularly evident when dealing with functions that exhibit strong nonlinearity or data with complex structures.

{Multi-Layer Perceptrons} (MLPs) \cite{cybenko1989approximation}, a type of neural networks, have been widely studied as a representative approach to function approximation. MLPs have been proven to approximate any continuous function with arbitrary precision using fixed nonlinear activation functions \cite{hornik1989multilayer}. Their effectiveness has been demonstrated in various fields, including modeling complex physical phenomena, image recognition, and recent large-scale language models \cite{de2023optimizing,preen2021autoencoding,narayanan2021efficient}. 
 However, their reliance on fixed nonlinear activation functions often requires large numbers of parameters to approximate complex functions effectively, which leads to inefficiencies \cite{mohan2024kans}.

{Kolmogorov-Arnold Networks} (KANs) \cite{liu2024kan}, inspired by the Kolmogorov-Arnold representation theorem \cite{kolmogorov1961representation,arnol1957functions}, have recently been proposed as a promising alternative. Unlike MLPs, KANs utilize spline-based learnable activation functions, enabling them to achieve higher parameter efficiency and scalability for irregular functions. KANs have shown superior performance in diverse applications such as image processing, time series prediction, and natural language processing \cite{cheon2024kolmogorov,somvanshi2024survey,livieris2024c}. Given their early successes, KANs are expected to attract increasing attention for a wide range of applications \cite{xu2024kolmogorov}.

Despite their advantages, KANs share a common limitation with MLPs: they rely on a single global model to approximate the entire problem space. This approach implicitly assumes that all data points follow the same underlying function. However, in reality, there are problems that cannot be adequately solved by a single global model. For instance, functions with local complexities (e.g., Fig \ref{fig: all functions}) or discontinuities (e.g., Fig. \ref{fig: ground truth}) pose significant challenges. Such problems are frequently encountered in real-world applications, including stock price prediction \cite{tang2019new} and multi-stage control system analysis \cite{gandomi2011multi}.

To address these concerns, the {X Classifier System for Function Approximation} (XCSF) \cite{wilson2002classifiers}, a widely studied evolutionary rule-based machine learning algorithm \cite{siddique2024survey}, may offer a promising solution. XCSF employs a divide-and-conquer approach that adaptively partitions the input space and assigns a distinct approximation model to each local region. XCSF enables localized modeling of complex functions, potentially complementing KAN's limitations in handling locally complex structures.

This paper proposes X-KAN, a function approximation method that simultaneously utilizes KAN's high representational power and XCSF's adaptive partitioning capability. X-KAN represents the entire input space using multiple local KAN models. Specifically, it defines a local region in the rule antecedent (i.e., IF part) and implements a KAN model in the rule consequent (i.e., THEN part), expressing local KAN models as rules. These IF-THEN rules are evolutionarily optimized by the XCSF framework. X-KAN is expected to improve approximation accuracy for functions with inherent local nonlinearities or discontinuities compared to a conventional single global KAN model.

The contributions of this paper are as follows:
\begin{itemize}
    \item We integrate KANs into evolutionary rule-based machine learning for the first time by introducing KANs into the XCSF framework. The effectiveness of this idea is demonstrated through artificial and real-world function approximation problems.

    \item We propose the first algorithm to automatically optimize multiple local KAN models. This results in a significant reduction in testing approximation error compared to conventional single global KAN models. We also explain that this effectiveness is due to XCSF's fundamental principle of assigning high fitness to local models (i.e., rules) with high generality and accuracy.
\end{itemize}

Note that recent studies have explored combining multiple KAN models for improved accuracy in various settings. For instance, Ensemble-KAN \cite{de2024ensemble} constructs several KANs using different subsets of input features and aggregates their outputs. Federated-KANs \cite{zeydan2025f} focus on distributed training across federated clients. Unlike these approaches, X-KAN uniquely integrates adaptive input space partitioning with local KAN optimization. Moreover, in contrast to approaches that predefine domain decompositions \cite{howard2024finite}, X-KAN simultaneously learns both the optimal partitioning of the input space and the parameters of local KAN models. This dual optimization enables X-KAN to automatically discover and model complex local structures and discontinuities in data, setting it apart from existing methods.

The remainder of this paper is organized as follows. Section \ref{sec: background} provides background on MLPs, KANs, and XCSF. Section \ref{sec: x-kan} presents our proposed algorithm, X-KAN. Section \ref{sec: experiments} reports and discusses experimental results. Section \ref{sec: further studies} presents further studies. Finally, Section \ref{sec: concluding remarks} concludes the paper.

\section{Background}
\label{sec: background}
\subsection{MLPs and KANs}
\subsubsection{Multi-Layer Perceptrons (MLPs)}
Multi-Layer Perceptrons (MLPs) are feedforward neural network architectures widely studied for function approximation problems. For a three-layer MLP with $n$ inputs, $H$ hidden nodes, and a single output, the forward computation can be expressed in matrix form as:
\begin{equation}
\label{eq: mlp}
    \operatorname{MLP}(\mathbf{x}) = \mathbf{w}^{(2)}\circ \sigma(\mathbf{W}^{(1)}\circ \mathbf{x} + \mathbf{b}^{(1)}) + b^{(2)},
\end{equation}
where $\mathbf{W}^{(1)} \in \mathbb{R}^{H \times n}$ is the weight matrix between the input and hidden layers, $\mathbf{w}^{(2)} \in \mathbb{R}^{1 \times H}$ is the weight vector between the hidden and output layers, $\mathbf{b}^{(1)} \in \mathbb{R}^H$ and $b^{(2)} \in \mathbb{R}$ are bias terms, and $\sigma$ is a fixed nonlinear activation function, e.g., Sigmoid-weighted Linear Unit (SiLU) \cite{elfwing2018sigmoid}. The MLP architecture is illustrated in Appendix \ref{sec: sup Network Architectures of MLPs and KANs}. According to \cite{yu2024kan}, the total number of parameters for this three-layer MLP, denoted as $N_\text{MLP}$, is:
\begin{equation}
\label{eq: number of parameters mlp}
    N_\text{MLP}=(nH+H)+(H+1)=H(n+2)+1.
\end{equation}

Based on the Universal Approximation Theorem (UAT) \cite{hornik1989multilayer}, MLPs with a single hidden layer can approximate any continuous function on compact subsets of $\mathbb{R}^n$ to arbitrary precision, as detailed in Appendix \ref{sec: sup Universal Approximation Theorem}. However, MLPs often require a large number of parameters to approximate complex functions  \cite{mohan2024kans}.

\subsubsection{Kolmogorov-Arnold Networks (KANs)}
Kolmogorov-Arnold Networks (KANs) are neural networks designed based on the Kolmogorov-Arnold representation theorem (KART). Further details of KART are provided in Appendix \ref{sec: sup Kolmogorov-Arnold Representation Theorem}. For a three-layer KAN with $n$ inputs, the matrix representation is:
\begin{equation}
\label{eq: kan}
    \operatorname{KAN}(\mathbf{x})=\boldsymbol{\phi}^{(2)}\circ \boldsymbol{\Phi}^{(1)}\circ \mathbf{x},
\end{equation}
where:
\begin{equation}
    \boldsymbol{\Phi}^{(1)} = \{\phi_{q,p}^{(1)} : [0,1] \to \mathbb{R} \mid p = 1, \ldots, 2n+1; q = 1, \ldots, n\}
\end{equation}
represents the first layer as a collection of learnable univariate activation functions, and:
\begin{equation}
    \boldsymbol{\phi}^{(2)} = \{\phi_{p}^{(2)} : \mathbb{R} \to \mathbb{R} \mid p = 1, \ldots, 2n+1\}
\end{equation}
represents the second layer as a collection of learnable univariate activation functions.
The KAN model expressed in Eq. \eqref{eq: kan} is mathematically equivalent to the KART formulation, which is presented in Eq. \eqref{eq: kart} of Appendix \ref{sec: sup Kolmogorov-Arnold Representation Theorem}.

Each activation function $\phi(x)$ is parameterized as:
\begin{equation}
    \phi(x) = w_b \cdot \operatorname{SiLU}(x) + w_s \cdot \operatorname{spline}(x),
\end{equation}
where $\operatorname{SiLU}(x) = x/(1 + e^{-x})$, $\operatorname{spline}(x)=\sum_{i=1}^{G+K}c_iB_i(x)$ with B-spline basis functions $B_i(x)$ \cite{de1978practical}, and $c_i$s, $w_b$, and $w_s$ are optimized via backpropagation. The KAN architecture is illustrated in Appendix \ref{sec: sup Network Architectures of MLPs and KANs}. According to \cite{yu2024kan}, the total number of parameters for this three-layer KAN, denoted as $N_\text{KAN}$, is:
\begin{equation}
\label{eq: number of parameters kan}
    N_\text{KAN}=(2n^2 + 3n + 1)(G + K) + (6n^2 + 11n + 5),
\end{equation}
where $G$ is the number of B-spline grids and $K$ is the B-spline degree.

Unlike MLPs, which use a fixed activation function at each node, KANs implement a learnable activation function on each edge between nodes. This design allows KANs to capture complex nonlinear relations more efficiently than MLPs. By leveraging KART, a KAN model handles the learning of a high-dimensional function as the learning of multiple univariate functions. This enables KANs to achieve higher parameter efficiency compared to MLPs, especially for problems involving complex data \cite{liu2024kan}. 
However, it is important to note that KART guarantees representations only for continuous functions. As a result, KAN may struggle to approximate discontinuous functions effectively.

\subsection{XCSF}
\subsubsection{Overview}
Wilson's X Classifier System for Function Approximation (XCSF)
\shortcite{wilson2002classifiers} is a rule-based piecewise function approximation method that implements linear regression models in rule consequents and hyperrectangular partitions in rule antecedents.\footnote{XCSF is a representative method of Learning Classifier Systems (LCS) \cite{urbanowicz2017introduction}, a family of rule-based machine learning algorithms. Historically, rules in LCS, including XCSF, are called \textit{classifiers} even when they work as regression models rather than classification models \cite{patzel2022bayesian}. To avoid this ambiguity in terminology, we use the term \textit{rule}.} XCSF combines evolutionary algorithms (EA) and stochastic gradient descent to generate general linear models with wide matching ranges in rule antecedents.

The key characteristics of XCSF are twofold. First, XCSF employs a subsumption operator \cite{wilson1998generalization} that aggregates multiple similar rules into a single, more general rule. Second, XCSF evaluates rule fitness based on both the number of aggregated rules and approximation error, optimizing the rule structure based on this fitness. Based on these two characteristics, XCSF promotes the acquisition of general linear models and realizes its fundamental principle of efficiently approximating data points with as few linear models as possible. 
Appendix \ref{sec: sup Model Architecture of XCSF} provides further details of XCSF.

\subsubsection{Extensions}
Since Wilson \shortcite{wilson2002classifiers} proposed XCSF, various extensions have been developed. For rule consequents, polynomial models \cite{lanzi2005extending}, MLPs \cite{lanzi2006xcsf}, support vector machines \cite{loiacono2007support}, and radial basis functions \cite{stein2018interpolation} have been proposed to enable the approximation of more complex nonlinear functions. For rule antecedents, hyperellipsoids \cite{butz2008function}, curved polytopes \cite{shiraishi2022beta}, convex hulls \cite{lanzi2006using}, 
gene expression programming \cite{wilson2006classifier}, and MLPs \cite{bull2002accuracy} have been introduced to achieve more flexible input space partitions. These extensions contribute to improving XCSF's approximation accuracy. Recently, XCSF was first applied to unsupervised autoencoding tasks by using MLPs as rule antecedents and autoencoders as rule consequents \cite{preen2021autoencoding}.

These extensions show XCSF's extensibility and validate XCSF's core principle of decomposing complex input spaces through rules. Our proposed integration of KAN into XCSF's rule consequents leverages KAN's universal approximation capabilities \cite{hecht1987kolmogorov}, guaranteed by KART, to achieve superior approximation accuracy compared to traditional XCSF approaches.

\section{X-KAN}
\label{sec: x-kan}
\begin{figure}[h]
\vspace{-5mm}
\centerline{\includegraphics[width=\linewidth]{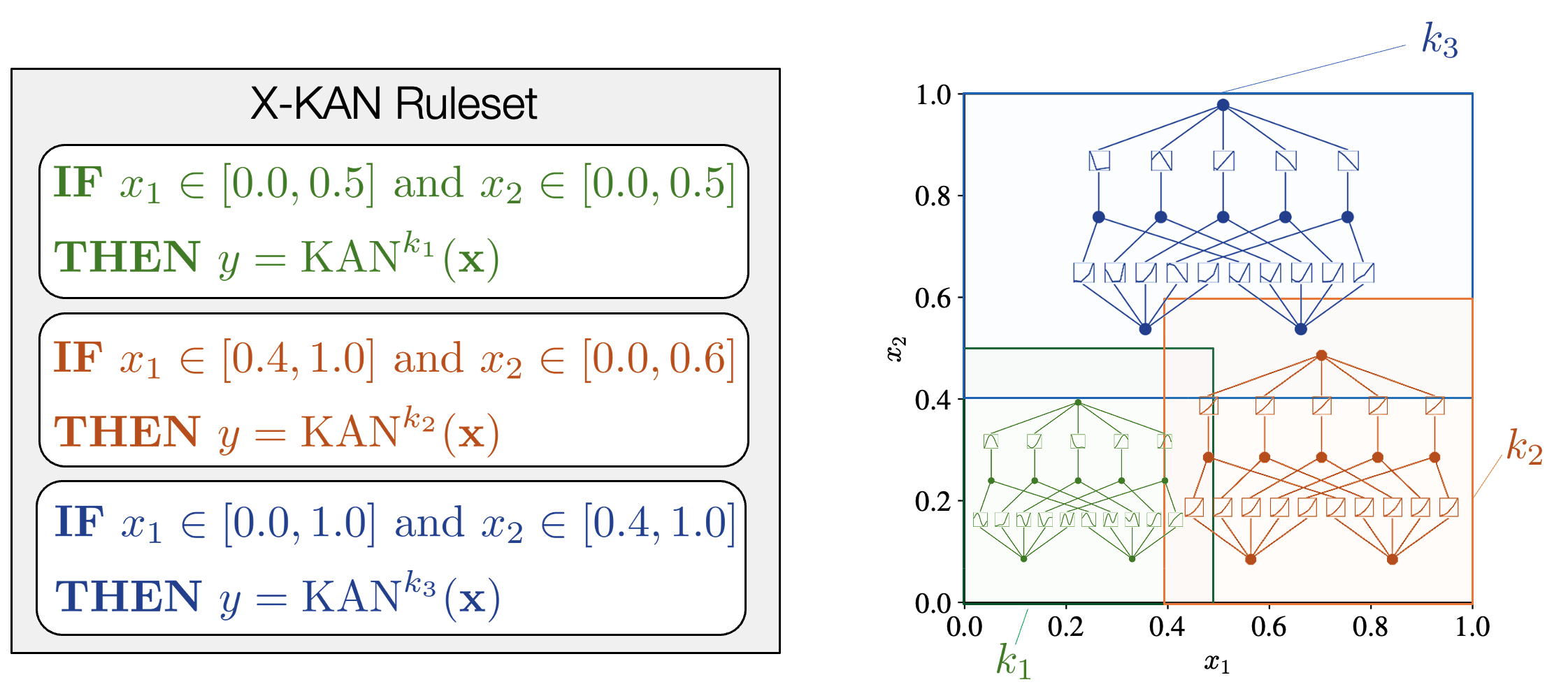}}
\vspace{-1mm}
\caption{An example of a ruleset of X-KAN with three rules, $k_1, k_2$, and $k_3$, in an input space $[0,1]^2$. X-KAN partitions the input space into local hyperrectangular regions defined by rule antecedents and performs local function approximation within each region using a KAN model implemented in the rule consequent. Appendix \ref{sec: sup the architecture of X-KAN} schematically illustrates the architecture of X-KAN.}
\label{fig: x-kan architecture}
\end{figure}
\vspace{-1mm}
We introduce X-KAN, a function approximation method that optimizes multiple local KAN models. Fig. \ref{fig: x-kan architecture} schematically illustrates X-KAN. X-KAN leverages the strengths of KAN's high expressiveness and XCSF's adaptive partitioning to address the limitations of conventional global function approximators. X-KAN has three key characteristics:
\begin{itemize}
    \item \emph{Local KAN Rules.} Each rule in X-KAN consists of an antecedent defined as an $n$-dimensional hyperrectangle and a consequent implemented as a single KAN model. This structure allows each rule to play a role as a local function approximator, activating only for the local region specified by its antecedent.

    \item \emph{Dual Optimization.} Utilizing the XCSF framework, X-KAN constructs general and accurate local KAN models with wide matching ranges. The rule antecedents are optimized through the EA, while the rule consequents (local KAN models) are optimized via backpropagation. This dual optimization enables X-KAN to adaptively place optimal local KAN models in each local region.

    \item \emph{Divide-and-Conquer.} Unlike KAN that operates as a single global function approximator, X-KAN functions as a divide-and-conquer algorithm by integrating multiple local KAN models having distinct activation functions.
\end{itemize}

\subsection{Rule Representation}
An $n$-dimensional rule $k$ in X-KAN is expressed as:
\begin{align}
\text{Rule }k:
\textbf{IF}\ x_1\in\left[l_1^k,u_1^k\right]\ \text{and}\ \ldots\ \text{and}\ x_n\in\left[l_n^k,u_n^k\right]\ \nonumber\\\textbf{THEN}\ y=\operatorname{KAN}^k(\mathbf{x})\ \textbf{WITH}\ F^k,
\end{align}
where:
\begin{itemize}
    \item $\mathbf{A}^k = (\mathbf{l}^k, \mathbf{u}^k) = (l_i^k,u_i^k)_{i=1}^n$ is the antecedent as a hyperrectangle 
    with bounds $\mathbf{l}^k, \mathbf{u}^k \in[0,1]^n$ and $l_i^k < u_i^k$ for all $i$ \cite{stone2003real};
    \item $\operatorname{KAN}^k(\cdot)$ is the consequent KAN model of rule $k$;
    \item $F^k \in (0,1]$ is the fitness value that evaluates both accuracy and generality of rule $k$.
\end{itemize}

Each rule $k$ maintains four key parameters:
(i) \textit{Error} $\epsilon^k \in \mathbb{R}_0^+$, representing absolute approximation error of the consequent KAN model;
(ii) \textit{Accuracy} $\kappa^k\in(0,1]$, calculated based on the error;
(iii) \textit{Generality} $\operatorname{num}^k\in \mathbb{N}_0$, indicating the number of aggregated rules;
and
(iv) \textit{Time stamp} $\operatorname{ts}^k \in \mathbb{N}_0$, representing the last time the rule was a candidate for the EA.

$\mathbf{A}^k$, $\operatorname{KAN}^k(\cdot)$, $\epsilon^k$, and $\kappa^k$ are uniquely determined and fixed when a rule is generated, either during the covering operation or by the EA. In contrast, $F^k$, $\operatorname{num}^k$, and $\operatorname{ts}^k$ are dynamically updated throughout the training process.

\subsection{Algorithm}
X-KAN operates in two distinct modes: training mode and testing mode. In training mode, X-KAN explores the search space to identify an accurate and general ruleset using a training dataset. Afterward, it performs rule compaction to produce a compact ruleset, denoted as $\pc$. In testing mode, $\pc$ is used to predict the output for testing data points.
\subsubsection{Training Mode}
\label{sss: x-kan training}
Algorithm \ref{alg: x-kan training} presents our algorithm for X-KAN training.
\begin{algorithm}[tb]
\footnotesize
    \caption{X-KAN training mode}
    \label{alg: x-kan training}
    \textbf{Input}: the training dataset $\mathcal{D}=(\mathcal{X},\mathcal{Y})=\{(\mathbf{x}, y) \in [0,1]^n \times \mathbb{R}\}$;\\
    \textbf{Output}: the compacted ruleset $\pc$;
    \begin{algorithmic}[1] 
        \STATE Initialize time $t$ as $t\leftarrow 0$;
        \STATE Initialize ruleset $\pop$ as $\pop\leftarrow \emptyset$;
        \WHILE{$t<\text{the maximum number of iterations}$}
        \STATE Observe a data point $(\mathbf{x},y)\in\mathcal{D}$;
        \STATE Create match set $\ms\subseteq\pop$ as in Eq. (\ref{eq: generate_match_set});
        \IF {$\ms=\emptyset$}
        \STATE Do covering to generate a new rule $k_c$;
        \STATE Insert $k_c$ to $\pop$ and $\ms$;
        \ENDIF
        \STATE Update $F^k$ for $\forall k\in\ms$ as in Eq. (\ref{eq: fitness});
        \IF {$t - \sum_{k\in{\ms}}{\rm num}^k \cdot {\rm ts}^k / {\sum_{k\in{\ms}}{\rm num}^k}>\theta_\text{EA}$}
        \STATE Update time stamp ${\rm ts}^k$ for $\forall k\in\ms$ as ${\rm ts}^k\leftarrow t$;
        \STATE Run EA on $\ms$;
        \STATE Do subsumption;
        \ENDIF
        \STATE Update time $t$ as $t \leftarrow t + 1$;
        \ENDWHILE
        \STATE Create compacted ruleset $\pc\subseteq\pop$ as in Eq. (\ref{eq: rule compaction});
        \STATE \textbf{return} $\pc$
    \end{algorithmic}
\end{algorithm}

\emph{Match Set Formation and Covering Operation.} Let $\pop$ be the current population of rules. At time $t$, a data point $(\mathbf{x}, y)$ is randomly sampled from the training dataset $\mathcal{D}$ (line 4). Subsequently, a match set $\ms$ is formed as (line 5):
\begin{equation}
\label{eq: generate_match_set}
    \ms = \{k \in \pop \mid \mathbf{x} \in \mathbf{A}^k\}.
\end{equation}
If $\ms = \emptyset$, a new rule $k_c$ satisfying $\mathbf{x} \in \mathbf{A}^{k_c}$ is generated and inserted into both $\pop$ and $\ms$ (lines 6–9). This operation is referred to as covering \cite{wilson2002classifiers}. Specifically, using hyperparameters $r_0 \in (0,1]$ and $P_\# \in [0,1]$, the antecedent of $k_c$, $\mathbf{A}^{k_c} = (\mathbf{l}^{k_c}, \mathbf{u}^{k_c})$, is determined as:
\begin{align}
\label{eq: covering}
\resizebox{\columnwidth}{!}{$
\left(l_i^{k_c},u_i^{k_c}\right)_{i=1}^n
=
\begin{cases}
(0,1) & \text{if}\ \mathcal{U}[0,1)<P_\#,\\
(x_i-\mathcal{U}(0,r_0], x_i+\mathcal{U}(0,r_0]) & \text{otherwise,}
\end{cases}
$}
\end{align}
where $\mathcal{U}(0, r_0]$ represents a random number uniformly sampled from the range $(0, r_0]$. Eq. \eqref{eq: covering} ensures that the range for $x_i$ is set to \textit{Don't Care} with probability $P_\#$, while otherwise it is set to a region encompassing $x_i$, based on the hyperparameter $r_0$. Next, the subset of data points within the range of $k_c$, denoted as $\mathcal{D}_{k_c}$, is constructed as:
\begin{equation}
\label{eq: data subset}
    \mathcal{D}_{k_c}=\{(\mathbf{x},y)\in\mathcal{D}\mid\mathbf{x}\in\mathbf{A}^{k_c}\}.
\end{equation}
Using this dataset, the local KAN model of $k_c$, $\operatorname{KAN}^{k_c}(\cdot)$, is trained via backpropagation for a specified number of epochs. The error of $k_c$, $\epsilon^{k_c}$, is then calculated as:
\begin{equation}
\label{eq: epsilon}
    \epsilon^{k_c}=\frac{1}{|\mathcal{D}_{k_c}|}\sum_{(\mathbf{x},y)\in\mathcal{D}_{k_c}}|y-\operatorname{KAN}^{k_c}(\mathbf{x})|.
\end{equation}
The error serves as the absolute error for the consequent KAN model of the rule. After that, the accuracy of $k_c$, $\kappa^{k_c}$, is calculated as:
\begin{equation}
\label{eq: absolute accuracy}
    \kappa^{k_c}=\begin{cases}
1 & \text{if}\ \epsilon^{k_c}<\epsilon_0,\\
\epsilon_0/\epsilon^{k_c}& \text{otherwise,}
\end{cases}
\end{equation}
where $\epsilon_0\in\mathbb{R}^+$ is a target error threshold (hyperparameter).
Subsequently, key parameters are initialized as follows: $F^{k_c}=0.01$, $\operatorname{num}^{k_c}=1$, and $\operatorname{ts}^{k_c}=0$.

\emph{Rule Fitness Update.} The fitness of each rule $k \in \ms$ is updated using the Widrow-Hoff learning rule \cite{widrow1960adaptive} as (line 10):
\begin{equation}
\label{eq: fitness}
    F^k\leftarrow F^k+\beta\left(\frac{\kappa^k\cdot\operatorname{num}^k}{\sum_{q\in\ms}\kappa^q\cdot\operatorname{num}^q}-F^k\right),
\end{equation}
where $\beta \in [0,1]$ is the learning rate. As indicated by Eq. \eqref{eq: absolute accuracy}, X-KAN (based on XCSF) defines a rule $k$ as \textit{accurate} when its approximation error satisfies $\epsilon^k < \epsilon_0$ (i.e., $\kappa^k = 1$). Consequently, the fitness $F^k$, as defined in Eq. \eqref{eq: fitness}, assigns higher values to rules with both smaller approximation errors $\epsilon^k$ (i.e., higher accuracy $\kappa^k$) and larger generality $\operatorname{num}^k$.

\emph{Application of the EA.} After updating the rules, the EA is applied to $\ms$ (lines 11–15). The EA is triggered when the average time since its last application over all rules in $\ms$ exceeds a threshold defined by the hyperparameter $\theta_{\mathrm{EA}}$. In this case, two parent rules $k_{p_1}$ and $k_{p_2}$ are selected from $\ms$ using tournament selection with a tournament size of $\tau$.
The selected parent rules are duplicated to create two offspring rules $k_{o_1}$ and $k_{o_2}$. Crossover is applied to their antecedents with a probability of $\chi$. During crossover, for each input dimension $i$, the lower bound $l_i$ and the upper bound $u_i$ are swapped between the two parents with a probability of 0.5 (i.e., uniform crossover). Subsequently, mutation is applied to each input dimension of the offspring with a probability of $\mu$. In mutation, a random value sampled from a uniform distribution, $\mathcal{U}[-m_0, m_0)$, is added to the bounds $l_i^{k_o}$ and $u_i^{k_o}$, where $k_o \in \{ k_{o_1}, k_{o_2} \}$ and $m_0\in\mathbb{R}^+$ is the maximum mutation magnitude.
If the resulting offspring rules $ k_o $ have antecedents that differ from their parents, their parameters are reinitialized as follows:
\begin{enumerate}
    \item Construct the subset of data points within the antecedent range of $ k_o $, denoted as $ \mathcal{D}_{k_o} $, using Eq. \eqref{eq: data subset}.
    \item Initialize $ \operatorname{KAN}^{k_o}(\cdot) $ and train it via backpropagation for a specified number of epochs.
    \item Calculate $ \epsilon^{k_o} $ and $ \kappa^{k_o} $ using Eqs. \eqref{eq: epsilon} and \eqref{eq: absolute accuracy}.
    \item Set $ F^{k_o} = 0.1 \cdot f $, where $ f = (F^{k_{p_1}} + F^{k_{p_2}})/2 $ if crossover occurs; otherwise, $ f = F^{k_o} $. Set $ \operatorname{num}^{k_o} = 1 $.
\end{enumerate}
 Finally, $k_{o_1}$ and $k_{o_2}$ are added to $\pop$ if they are not subsumed by their parent rules (described below). If the total generality in $\pop$,$\sum_{k\in\pop}\operatorname{num}^k$, exceeds the maximum ruleset size $N$, two rules are deleted, as in \cite{preen2021autoencoding}.

 \emph{Subsumption Operator.} X-KAN employs a subsumption operator \cite{wilson1998generalization} to aggregate offspring rules into more general parent rules (line 14). Specifically, for a parent rule $ k_p \in \{ k_{p_1}, k_{p_2} \} $ and an offspring rule $ k_o \in \{ k_{o_1}, k_{o_2} \} $:
 \begin{enumerate}
     \item The parent rule must be more general than the offspring rule (i.e., $ \mathbf{A}^{k_p} \supseteq \mathbf{A}^{k_o} $).
     \item The parent rule must be accurate (i.e., $ \kappa^{k_p} = 1$).
 \end{enumerate}
 If these conditions are met, the generality of the parent rule is updated as $\operatorname{num}^{k_p} \leftarrow
\operatorname{num}^{k_p} + 
\operatorname{num}^{k_o}$
 and the offspring rule $ k_o $ is removed from $\pop$.

 \emph{Rule Compaction.} After the training is completed, the rule compaction algorithm \cite{orriols2008fuzzy} is applied to obtain a compacted ruleset, denoted as $\pc$ (line 18). The compacted ruleset is defined as:
 \begin{equation}
\label{eq: rule compaction}
\pc = \bigcup_{\mathbf{x}\in\mathcal{D}} \left\{\arg\max_{k\in\ms} F^k\right\}.
\end{equation}
For each training data point $\mathbf{x}$, only the rule with the highest fitness (called \textit{single winner} rule) in its match set $
\ms = \{k \in \pop \mid \mathbf{x} \in \mathbf{A}^k\}$ is copied to $\pc$. 
Appendix \ref{ss: sup Effects of Rule Compaction on X-KAN} shows that the compaction enables X-KAN to reduce the number of rules by up to 72\% while maintaining approximation accuracy.
\subsubsection{Testing Mode}
For a testing data point $\mathbf{x}_\text{te}$, X-KAN computes the predicted value $\hat{y}_\text{te}$ using $\pc$, obtained during the training mode, and the single winner-based inference scheme \cite{ishibuchi1999performance}. The prediction is calculated as:
\begin{equation}
    \hat{{y}_\text{te}}=\operatorname{KAN}^{k^*}(\mathbf{x}_\text{te}),\quad \text{where}\quad k^*=\arg \max_{k\in\ms_\text{te}}(F^k),
\end{equation}
with $\ms_\text{te} = \{k \in \pc \mid \mathbf{x}_\text{te} \in \mathbf{A}^k\}$ as the testing match set and $k^*$ as the single winner rule. This inference ensures that only the rule with the highest fitness, which reflects both accuracy and generality, contributes to the prediction.

\section{Experiments}
\label{sec: experiments}
\subsection{Experimental Setup}
\label{ss: experimental setup}
We evaluate X-KAN’s performance on eight function approximation problems: four test functions shown in Fig. \ref{fig: all functions} from \cite{stein2018interpolation} and four real-world datasets from \cite{heider2023suprb}. For details of these problems, kindly refer to Appendices \ref{sec: sup description of the test functions} and \ref{sec: sup description of the real-world}. For each test function, we uniformly sample 1,000 data points to create a dataset.

We compare X-KAN against three baseline methods: XCSF, MLP, and KAN. Through the comparison with XCSF, we validate the effectiveness of extending rule consequents from linear models to KAN models. 
The comparison with MLP, a standard baseline in machine learning, allows us to evaluate X-KAN's overall performance. 
Finally, comparing X-KAN with KAN enables us to examine the benefits of extending from a single global model to multiple local models.

The hyperparameters for XCSF and X-KAN are set to $r_0=1.0$, $P_\#\in\{0.0\, \text{(test functions)},0.8\,\text{(real-world datasets)}\}$, $\epsilon_0=0.02$, $\beta=0.2$, $\theta_\text{EA}=100$, $\tau=0.4$, $\chi=0.8$, $\mu=0.04$, $m_0=0.1$, and $N=50$. The maximum number of training iterations for XCSF and X-KAN is 10 epochs. The same architecture in Eq. \eqref{eq: kan} is used for KAN and each rule in X-KAN, which consists of three layers with $2n+1$ nodes in the hidden layer, where $G=3$ and $K=3$ for B-spline parameters. The three-layer MLP architecture in Eq. \eqref{eq: mlp} with $H$ hidden nodes is used together with SiLU activation functions. For a fair comparison, $H$ is set for each problem such that the total number of parameters in MLP ($N_\text{MLP}$ in Eq. \eqref{eq: number of parameters mlp}) equals that of KAN ($N_\text{KAN}$ in Eq. \eqref{eq: number of parameters kan})\footnote{For example, in a two-dimensional input problem ($n=2$), KAN has $(2\cdot2^2 + 3\cdot2 + 1)(3 + 3) + (6\cdot2^2 + 11\cdot2 + 5) = 141$ parameters. Therefore, $H$ is set to 35 for MLP to match this parameter count, as $35\cdot(2+2)+1=141$.}. All network hyperparameters follow the original KAN authors' implementation\footnote{\url{https://github.com/KindXiaoming/pykan}}, with training conducted for 10 epochs.
Input features are normalized to $[0,1]$, and data targets are normalized to $[-1,1]$.

Performance evaluation uses Mean Absolute Error (MAE) on test data over 30 trials of Monte Carlo cross-validation, with 90\% training and 10\% testing data splits. 
Statistical significance is assessed through Wilcoxon signed-rank tests ($\alpha=0.05$) for each problem, while overall performance is compared using Friedman tests with Holm corrections, reporting both raw and Holm-adjusted $p$-values.

\subsection{Results}
\label{ss: results}
\begin{figure*}
\vspace{-10mm}
    \centering
    \begin{subfigure}[b]{0.24\textwidth}
        \centering
        \includegraphics[width=\textwidth]{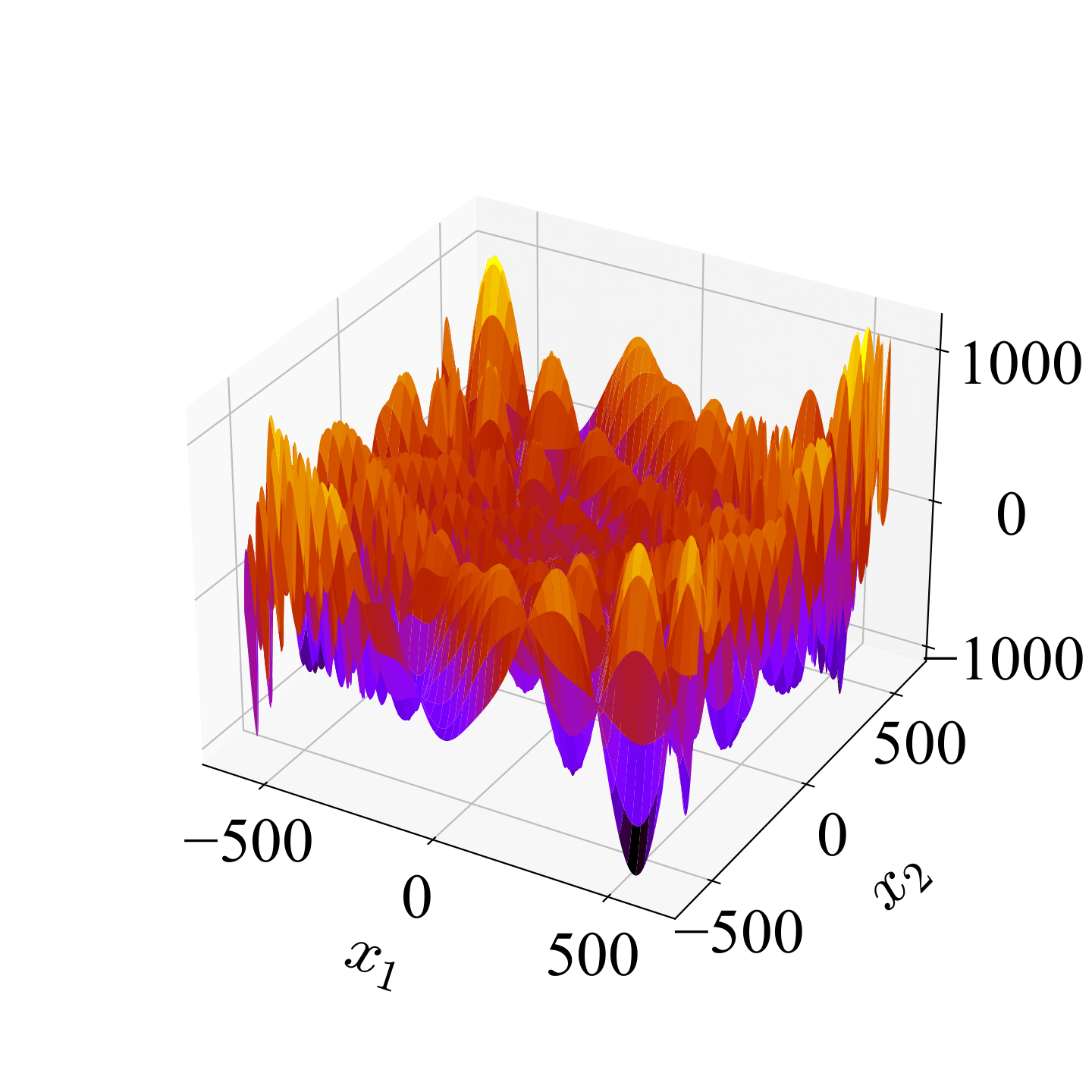}
        \vspace{-3mm}
        \caption{$f_1$: Eggholder function}
        \label{fig:egg}
    \end{subfigure}
    \hfill
    \begin{subfigure}[b]{0.24\textwidth}
        \centering
        \includegraphics[width=\textwidth]{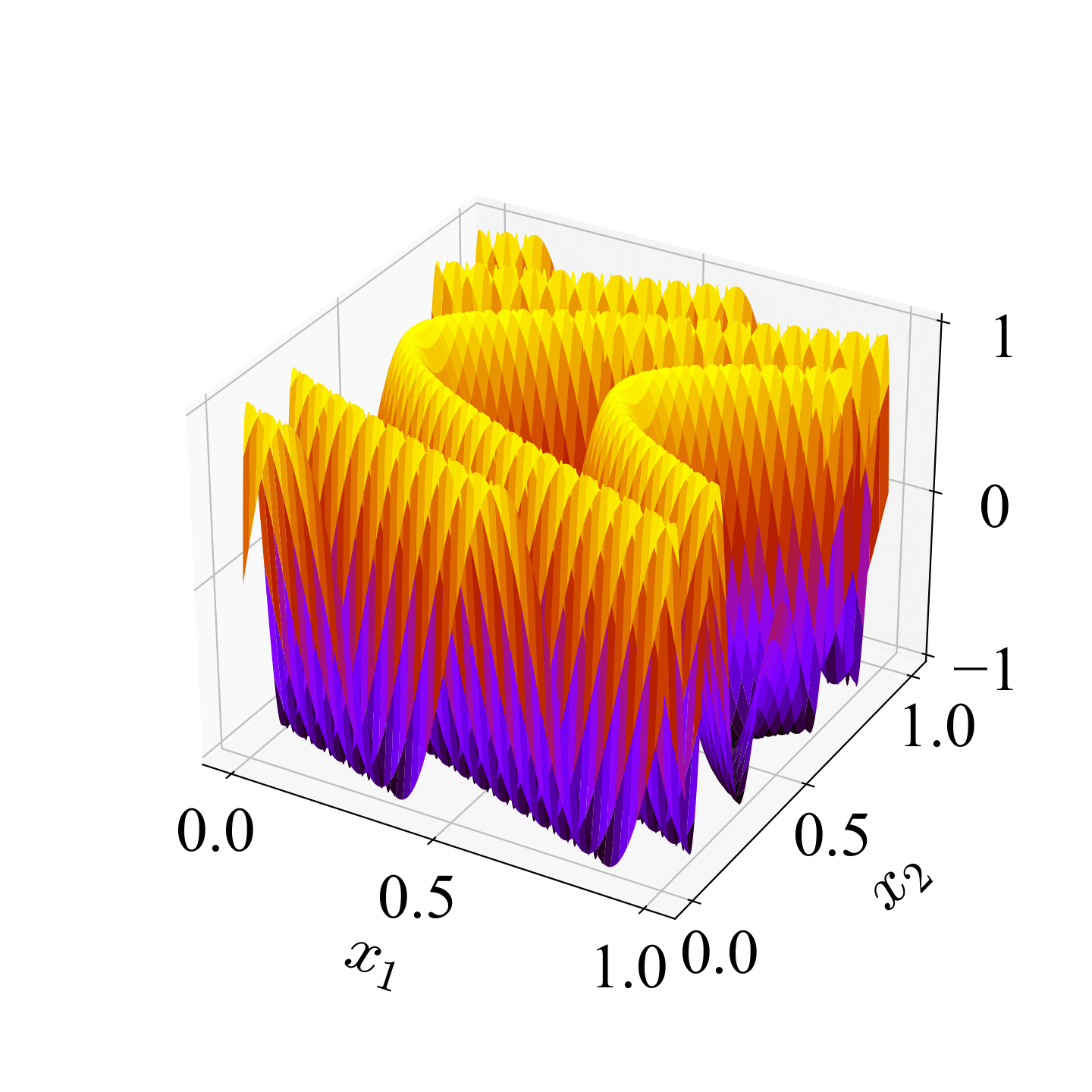}
        \vspace{-3mm}
        \caption{$f_2$: Sine-in-Sine function}
        \label{fig:sine}
    \end{subfigure}
    \hfill
    \begin{subfigure}[b]{0.24\textwidth}
        \centering
        \includegraphics[width=\textwidth]{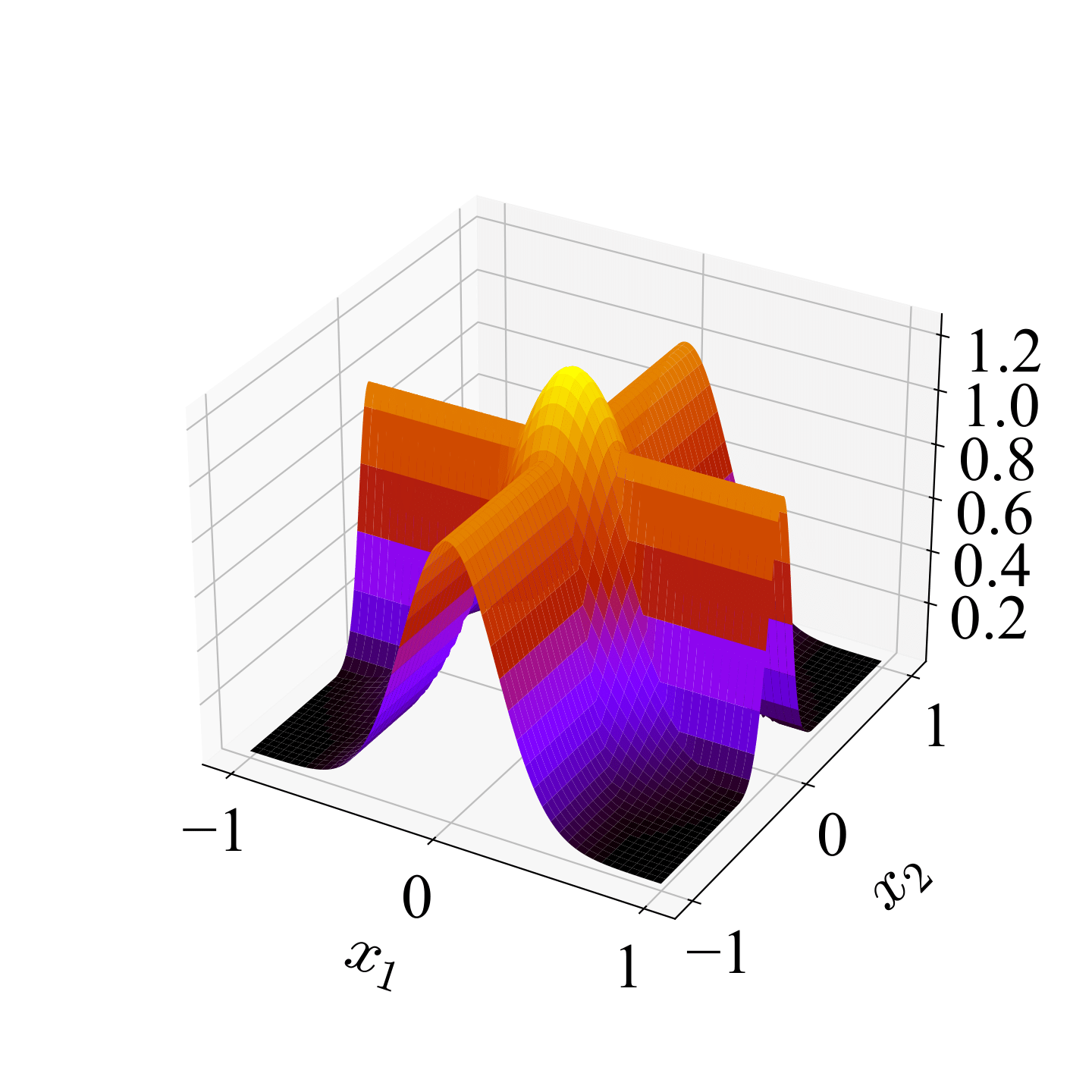}
        \vspace{-3mm}
        \caption{$f_3$: Cross function}
        \label{fig:cross}
    \end{subfigure}
    \hfill
    \begin{subfigure}[b]{0.24\textwidth}
        \centering
        \includegraphics[width=\textwidth]{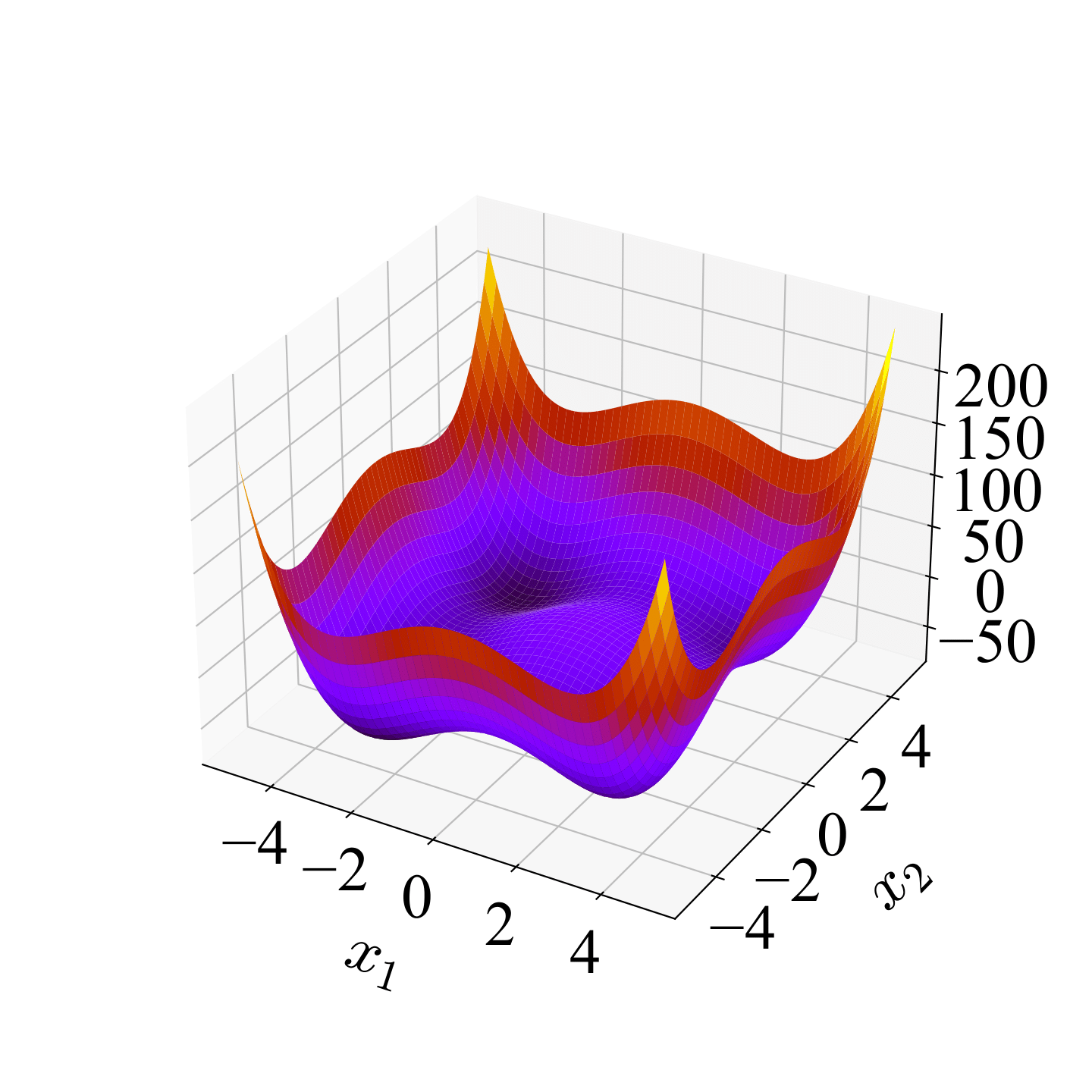}
        \vspace{-3mm}
        \caption{$f_4$: Styblinski-Tang function}
        \label{fig:styblinski}
    \end{subfigure}
    \vspace{-2mm}
    \caption{The two-dimensional test functions $f_1$-$f_4$. (a) $f_1$ has high curvature and non-linearity in both dimensions. (b) $f_2$ features intricate nonlinear patterns and multiple local extrema. (c) $f_3$ combines both linear and nonlinear local regions. (d) $f_4$ is characterized by relatively smooth regions in its central domain but exhibits steep gradients near its boundaries.}
    \label{fig: all functions}
\centering

    \vspace{5mm}
    \resizebox{\width}{!}{
\small
    \begin{tabular}{c l c c|cccc|cc}

\bhline{1pt}
\multicolumn{4}{l|}{8 \textsc{Datasets}} & \multicolumn{4}{c|}{\textsc{Testing Mean Absolute Error (MAE)}} & \multicolumn{2}{c}{\textsc{\#Rules in $\pc$}}\\
Abbr. & Name & \#Inst. & $n$ & XCSF & MLP & KAN & X-KAN & XCSF & X-KAN \\
 \bhline{1pt}
$f_1$ & Eggholder Function & 1000 & 2 & \cellcolor{p}0.27970 $-$ & 0.27170 $-$ & 0.22220 $-$ & \cellcolor{g}0.16970 & \cellcolor{g}5.600 $+$ & \cellcolor{p}9.200 \\
$f_2$ & Sine-in-Sine Function & 1000 & 2 & \cellcolor{p}0.64870 $-$ & 0.63820 $-$ & 0.44610 $-$ & \cellcolor{g}0.12730 & \cellcolor{g}4.600 $+$ & \cellcolor{p}10.50 \\
$f_3$ & Cross Function & 1000 & 2 & \cellcolor{p}0.33190 $-$ & 0.30570 $-$ & 0.06662 $-$ & \cellcolor{g}0.02379 & \cellcolor{p}6.767 $-$ & \cellcolor{g}4.867 \\
$f_4$ & Styblinski-Tang Function & 1000 & 2 & 0.18070 $-$ & \cellcolor{p}0.24100 $-$ & 0.12850 $-$ & \cellcolor{g}0.06922 & \cellcolor{g}4.833 $+$ & \cellcolor{p}6.400 \\
ASN & Airfoil Self-Noise & 1503 & 5 & 0.19240 $-$ & \cellcolor{p}0.19990 $-$ & 0.08407 $-$ & \cellcolor{g}0.05533 & \cellcolor{g}5.633 $+$ & \cellcolor{p}7.900 \\
CCPP & Combined Cycle Power Plant & 9548 & 4 & 0.09288 $-$ & \cellcolor{p}0.09521 $-$ & 0.08684 $-$ & \cellcolor{g}0.07871 & \cellcolor{p}8.567 $\sim$ & \cellcolor{g}7.800 \\
CS & Concrete Strength & 1030 & 8 & \cellcolor{p}0.19050 $-$ & 0.16490 $-$ & 0.08833 $-$ & \cellcolor{g}0.07842 & \cellcolor{g}7.233 $\sim$ & \cellcolor{p}7.900 \\
EEC & Energy Efficiency Cooling & 768 & 8 & \cellcolor{p}0.12560 $-$ & 0.12290 $-$ & 0.05232 $-$ & \cellcolor{g}0.02729 & \cellcolor{p}6.233 $-$ & \cellcolor{g}2.667 \\
\bhline{1pt}
\multicolumn{4}{r|}{Rank}& \cellcolor{p}\textit{3.62}$\downarrow^{\dag\dag}$ & \textit{3.38}$\downarrow^{\dag\dag}$ & \textit{2.00}$\downarrow^{\dag\dag}$ & \cellcolor{g}\textit{1.00} & \cellcolor{g}\textit{1.38}$\uparrow$ & \cellcolor{p}\textit{1.62} \\
\multicolumn{4}{r|}{Position}& \textit{4} & \textit{3} & \textit{2} & \textit{1} & \textit{1} & \textit{2} \\
\multicolumn{4}{r|}{Number of $+/-/\sim$} &  0/8/0 & 0/8/0 & 0/8/0 & - & 4/2/2 & - \\
\bhline{1pt}
\multicolumn{4}{r|}{$p$-value} & 0.00781 & 0.00781 & 0.00781 & - & 0.461 & - \\
\multicolumn{4}{r|}{$p_\text{Holm}$-value} & 0.0234 & 0.0234 & 0.0234 & - & - & - \\

\bhline{1pt}

\end{tabular}

}
\captionsetup{type=table}  
\caption{Summary of dataset characteristics and experimental results, displaying dataset information (\#Inst.: number of instances, $n$: number of inputs) and performance metrics (testing MAE and number of rules in $\pc$). Green/peach highlighting indicates best/worst values. Rank and Position indicate the average ranking from the Friedman test and final position, respectively. Symbols $+/-/\sim$ denote statistically significant better/worse/similar performance compared to X-KAN based on Wilcoxon signed-rank tests. Arrows $\uparrow/\downarrow$ indicate improvement/decline in rank compared to X-KAN. Statistical significance at $\alpha=0.05$ is denoted by $\dag$ ($p$-value) and $\dag\dag$ (Holm-adjusted $p$-value).}
    \label{tb:result}
 \captionsetup{type=figure}  
 \vspace{3mm}
     \begin{subfigure}[b]{0.24\textwidth}
        \centering
        \includegraphics[width=\textwidth]{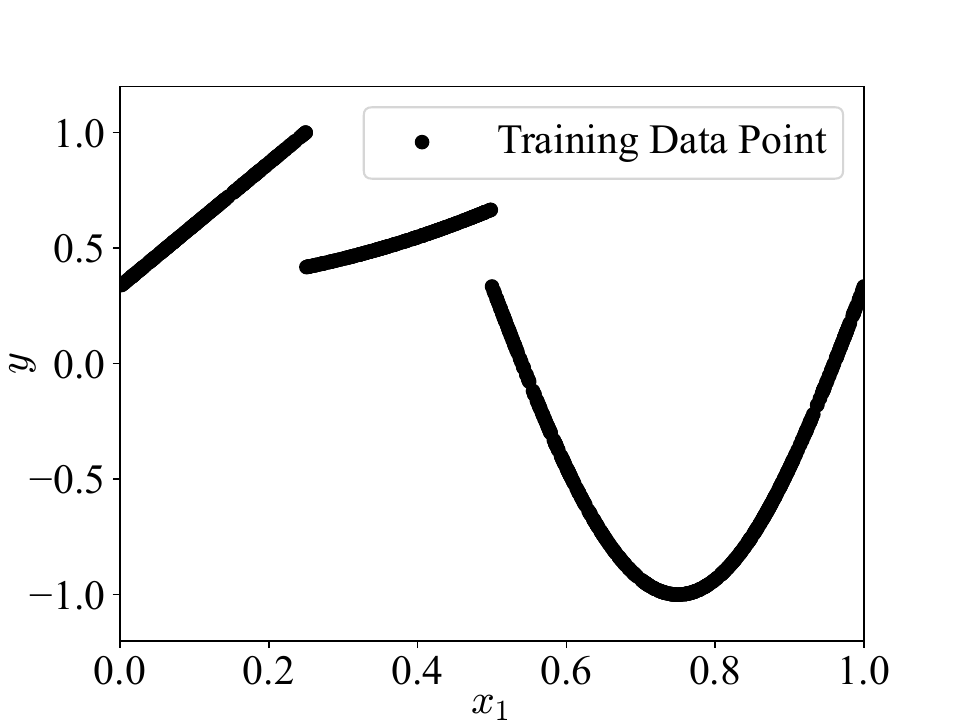}
        \caption{Training dataset}
        \label{fig: ground truth}
    \end{subfigure}
    \hfill
    \begin{subfigure}[b]{0.24\textwidth}
        \centering
        \includegraphics[width=\textwidth]{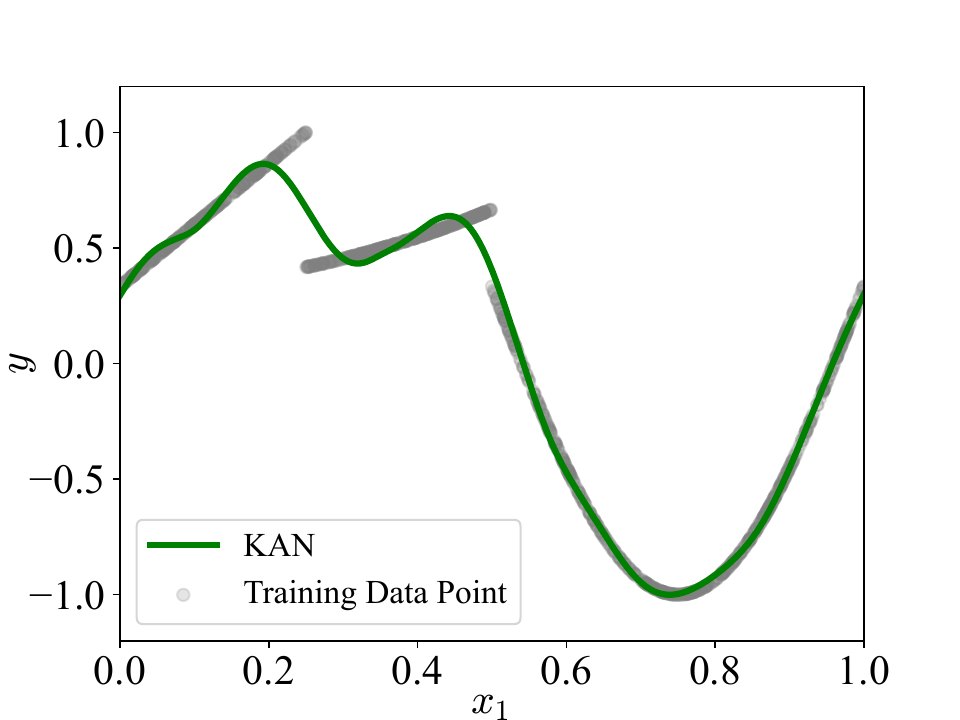}
        \caption{Prediction results of KAN}
        \label{fig: kan prediction}
    \end{subfigure}
    \hfill
    \begin{subfigure}[b]{0.24\textwidth}
        \centering
        \includegraphics[width=\textwidth]{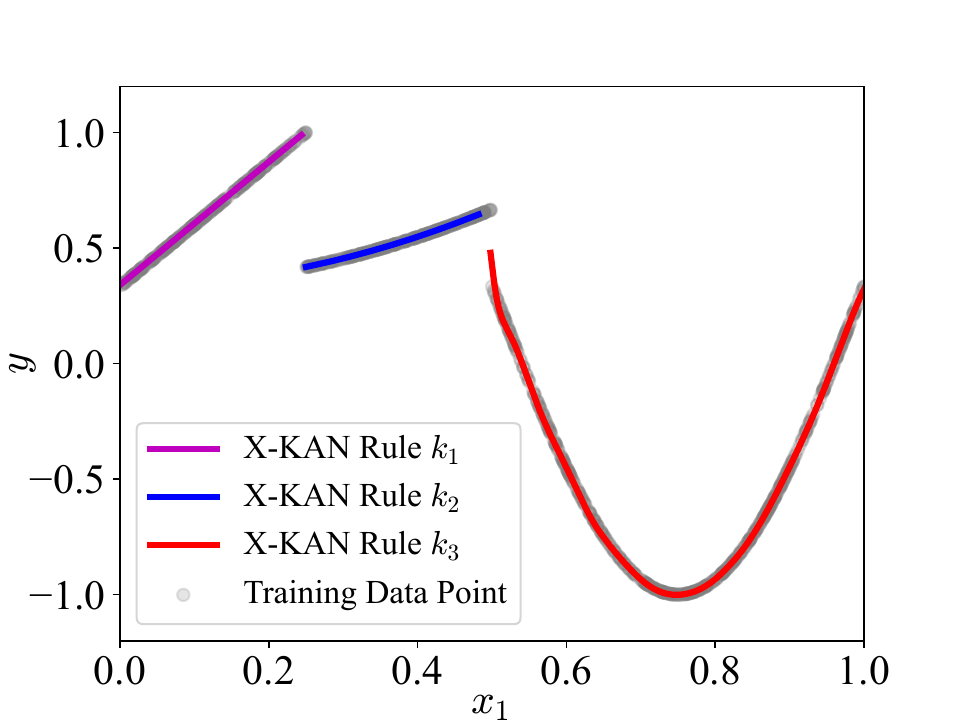}
        \caption{Prediction results of X-KAN}
        \label{fig: xkan prediction}
    \end{subfigure}
    \hfill
    \begin{subfigure}[b]{0.24\textwidth}
        \centering
        \includegraphics[width=\textwidth]{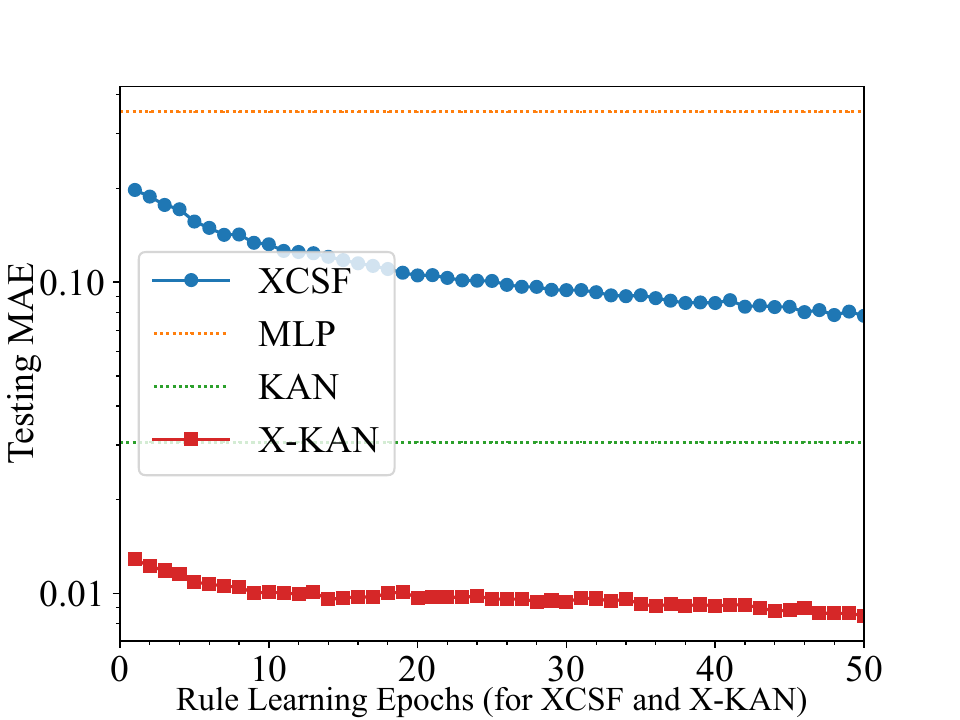}
        \caption{Testing MAE trends}
        \label{fig: syserr}
    \end{subfigure}
    \vspace{-2mm}
    \caption{Performance analysis of each method on a discontinuous function. (c) shows that X-KAN generated three rules.}
    \label{fig: discontinuous}

\vspace{5mm}
\captionsetup{type=table}  
 \begin{minipage}[b]{0.58\textwidth}
        
            \resizebox{0.86\width}{!}{
\small
\centering
\begin{tabular}{c |c c|cc|cc}

\bhline{1pt}
 & \multicolumn{2}{c|}{\textsc{Training MAE}} & \multicolumn{2}{c|}{\textsc{Testing MAE}} & \multicolumn{2}{c}{\textsc{\#Rules in $\pc$}}\\
  & X-KAN & X-KAN$\kappa$  & X-KAN & X-KAN$\kappa$ & X-KAN & X-KAN$\kappa$ \\
 \bhline{1pt}

$f_1$ & \cellcolor{p}0.10780  & \cellcolor{g}0.08943 $+$ & \cellcolor{p}0.16970 & \cellcolor{g}0.16880 $\sim$ & \cellcolor{g}9.200 & \cellcolor{p}18.60 $-$ \\
$f_2$ & \cellcolor{g}0.07356  & \cellcolor{p}0.07860 $\sim$&  \cellcolor{g}0.12730  & \cellcolor{p}0.12940 $\sim$& \cellcolor{g}10.50  & \cellcolor{p}16.30 $-$\\
$f_3$ &  \cellcolor{g}0.01927 & \cellcolor{p}0.01949 $\sim$ &  \cellcolor{g}0.02379  & \cellcolor{p}0.02397 $\sim$& \cellcolor{g}4.867  & \cellcolor{p}9.633 $-$ \\
$f_4$ &  \cellcolor{p}0.05171  & \cellcolor{g}0.04586 $\sim$&  \cellcolor{p}0.06922  & \cellcolor{g}0.06070 $\sim$& \cellcolor{g}6.400  & \cellcolor{p}11.83 $-$ \\
ASN &  \cellcolor{p}0.03548  & \cellcolor{g}0.03073 $+$ &  \cellcolor{p}0.05533  & \cellcolor{g}0.05461 $\sim$& \cellcolor{g}7.900  & \cellcolor{p}19.97 $-$\\
CCPP &  \cellcolor{g}0.05691  & \cellcolor{p}0.05971 $\sim$&  \cellcolor{g}0.07871  & \cellcolor{p}0.08174 $-$& \cellcolor{g}7.800  & \cellcolor{p}17.53 $-$\\
CS &  \cellcolor{p}0.02687  & \cellcolor{g}0.02497 $\sim$&  \cellcolor{g}0.07842  & \cellcolor{p}0.08014 $\sim$& \cellcolor{g}7.900 & \cellcolor{p}17.40 $-$ \\
EEC &  \cellcolor{p}0.01561  & \cellcolor{g}0.01490 $\sim$&  \cellcolor{g}0.02729  & \cellcolor{p}0.03004 $-$& \cellcolor{g}2.667  & \cellcolor{p}7.367 $-$\\
\bhline{1pt}
Rank  & \cellcolor{p}\textit{1.62} & \cellcolor{g}\textit{1.38}$\uparrow$ & \cellcolor{g}\textit{1.38} & \cellcolor{p}\textit{1.62}$\downarrow$ & \cellcolor{g}\textit{1.00} & \cellcolor{p}\textit{2.00}$\downarrow^{\dag}$ \\
Position & \textit{2} & \textit{1} & \textit{1} & \textit{2} & \textit{1} & \textit{2} \\
$+/-/\sim$ &  - & 2/0/6 &  - & 0/2/6 & - & 0/8/0 \\
\bhline{1pt}
$p$-value &  - & 0.383 &  - & 0.547 & - & 0.00781 \\

\hline
\bhline{1pt}

\end{tabular}
}
\captionsetup{type=table}  
\caption{Comparison of fitness functions: X-KAN (accuracy and generality)\\ vs. X-KAN$\kappa$ (accuracy only). Notation follows Table \ref{tb:result}.}
    \label{tb: x-kan kappa}
    \end{minipage}
\begin{minipage}[b]{0.38\textwidth}
        \centering
            \begin{tikzpicture}
  \begin{axis}[
    ybar,  
    bar width=10pt,
    ymin=0.1, 
    ymax=200,
    ymode=log,
    log origin=infty, 
    ylabel={Runtime (sec.)},
    xtick=data,
    symbolic x coords={XCSF, MLP, KAN, X-KAN},
    xticklabels={XCSF$_\texttt{.jl}$,MLP$_\texttt{.py}$,KAN$_\texttt{.py}$, X-KAN$_\texttt{.jl+.py}$},
     xticklabel style={rotate=90,anchor=east},
    nodes near coords,
    nodes near coords style={font=\scriptsize},
    nodes near coords style={
        /pgf/number format/.cd,
        fixed,
        precision=3,
    },
    point meta=rawy,  
    width=0.85\textwidth,
    height=0.54\textwidth
  ]
  \addplot coordinates { (XCSF, 0.13) (MLP, 1.24) (KAN, 6.01) (X-KAN, 42.9) };
  \end{axis}
\end{tikzpicture}
\captionsetup{type=figure} 
\caption{Average runtime per trial. Extensions \texttt{.jl}/\texttt{.py} indicate implementations in Julia/Python.}
    \label{fig: runtime}
    \end{minipage}

\end{figure*}

Table \ref{tb:result} presents the testing MAE of each method and the number of rules in the compacted ruleset $\pc$ of XCSF and X-KAN at the end of training. For XCSF and X-KAN, the MAE values are calculated using their compacted rulesets. 

The results show that X-KAN achieves significantly lower testing MAE than all baseline methods across all problems, with statistical significance confirmed by Holm-adjusted $p$-values ($p_\text{Holm} < 0.05$). 
Regarding the number of rules, XCSF generates 6.2 rules on average while X-KAN generates 7.2 rules. Although X-KAN generates slightly more rules than XCSF, this difference is not statistically significant ($p = 0.461$).
Appendix \ref{ss: sup training mae} shows that X-KAN's training MAE is also significantly lower than those of the baseline methods.
These findings demonstrate that X-KAN outperforms XCSF, MLP, and KAN in function approximation accuracy while maintaining compact rulesets.

Note that during inference, X-KAN uses only a single rule with an identical parameter count to the (global) KAN, ensuring a fair comparison of their core approximation capabilities. Appendix \ref{ss: comparison with x-mlp} shows that X-MLP, which replaces KANs with MLPs in our framework, outperforms the global MLP but is still outperformed by X-KAN. Appendix \ref{ss: comparison with widekan} shows that X-KAN significantly outperforms WideKAN, which expands the hidden layer to match X-KAN's total parameter count. These results confirm that X-KAN's advantage stems from its evolutionary search mechanism enabling specialized local approximations, rather than merely from increased parameters---a benefit that cannot be achieved simply by increasing the size of a single global model like WideKAN.

\subsection{Discussion}
As shown in Fig. \ref{fig: all functions}, $f_1$ and $f_2$ exhibit strong nonlinearity, making them challenging for XCSF's linear models to approximate effectively. This is reflected in Table \ref{tb:result}, where XCSF shows the highest MAE for these problems (highlighted in peach). Conversely, XCSF's linear models perform well on problems with mixed linear and nonlinear characteristics ($f_4$) and predominantly linear problems (CCPP \cite{heider2023suprb}), even outperforming MLP. These results align with previous findings that XCSF's performance strongly depends on problem linearity \cite{lanzi2007generalization}.

X-KAN significantly outperforms the compared algorithms for all problems. Most notably, on $f_2$, which features strong input interdependencies and high curvature, X-KAN shows large improvement over KAN. Regarding the number of rules, X-KAN generates more rules (9--10) for highly nonlinear problems ($f_1$, $f_2$) and fewer rules (4--6) for problems with lower interdependency and curvature ($f_3$, $f_4$), demonstrating its ability to adjust to problem complexity.

\section{Further Studies}
\label{sec: further studies}
\subsection{Analysis on a Discontinuous Function}
\label{ss: Analysis on a Discontinuous Function}
Since KAN is based on KART, which is designed for continuous functions, it may struggle to approximate discontinuous functions effectively. 
To validate this hypothesis, we conducted experiments using 1,000 data points samples from a discontinuous function with jump discontinuity used in \cite{shoji2023piecewise}, as shown in Fig. \ref{fig: ground truth}. For details of the function, kindly refer to Appendix \ref{sec: sup description of the discontinuous function}. The experimental settings followed Section \ref{ss: experimental setup}, except for $P_\# = 0.0$, $r_0 = 0.5$, $\epsilon_0=0.01$, maximum training iterations of 50 epochs for XCSF and X-KAN, and 50 epochs for MLP, KAN, and X-KAN's local KAN models. Figs. \ref{fig: kan prediction} and \ref{fig: xkan prediction} show the prediction plots from the best trials of KAN and X-KAN, respectively. Fig. \ref{fig: syserr} illustrates the decrease of the testing MAE during rule learning for XCSF and X-KAN, with the final MAE values for MLP and KAN (dashed horizontal lines).

Fig. \ref{fig: kan prediction} demonstrates that KAN fails to detect discontinuities, instead producing a smooth continuous function approximation. In contrast, Fig. \ref{fig: xkan prediction} shows that X-KAN successfully identifies discontinuities and performs piecewise function approximation using three rules. The decreasing MAE trends by XCSF and X-KAN in Fig. \ref{fig: syserr} validate the effectiveness of local approximation.

These results demonstrate X-KAN's ability to handle discontinuous functions through its adaptive partitioning approach, overcoming a fundamental limitation of KAN.

\subsection{Role of Generality in Fitness}
In Table \ref{tb:result}, X-KAN never shows higher testing MAE than KAN. This high performance can be attributed to X-KAN (XCSF)'s fundamental principle of assigning fitness based on both accuracy $\kappa^k$ and generality $\operatorname{num}^k$.
To validate this hypothesis, we conducted experiments comparing X-KAN against its variant that assigns fitness solely based on accuracy (denoted as X-KAN$\kappa$). For X-KAN$\kappa$, the fitness update rule was simplified to $F^k\leftarrow\kappa^k$. The experimental settings followed Section \ref{ss: experimental setup}.

In Table \ref{tb: x-kan kappa}, X-KAN$\kappa$ achieved significantly lower training MAE than X-KAN on two problems ($f_1$, ASN). However, for testing MAE, X-KAN$\kappa$ performed significantly worse than X-KAN on two problems (CCPP, EEC). This performance degradation can be attributed to the increased probability of selecting parent rules with high accuracy but low generality (overfitting to training data) when generality is not considered in fitness calculation. Consequently, X-KAN$\kappa$ generated approximately twice as many rules as X-KAN due to reduced generalization pressure.

These findings demonstrate that considering both accuracy and generality, as implemented in XCSF and X-KAN, is crucial for improving generalization performance in evolutionary rule-based machine learning models.

\subsection{Runtime Analysis}
Fig. \ref{fig: runtime} shows the average runtime per trial under an experimental environment running Ubuntu 24.04.1 LTS with an Intel® Core™ i9-13900F CPU (5.60 GHz) and 32GB RAM. While MLP, KAN, and local KAN models in X-KAN were implemented in Python by the original KAN authors, the XCSF and X-KAN frameworks were implemented in Julia \cite{bezanson2017julia} by the authors. X-KAN calls Python-based local KAN models from its Julia framework. 
Note that, due to the mixed-use of programming languages, the runtime comparisons should be interpreted with caution.

As shown in Fig. \ref{fig: runtime}, KAN requires approximately five times more runtime than MLP, mainly due to the computational overhead of recursive B-spline functions \cite{qiu2024powermlp}. Furthermore, X-KAN requires approximately seven times more runtime than KAN, as each rule's consequent implements a separate KAN model that must be trained.

One idea to decrease the runtime of X-KAN is to decrease the number of rules by increasing the generality of each rule. For example, dynamic adjustment of the target error threshold $\epsilon_0$ used in subsumption operations \cite{hansmeier2020adaption} can reduce the total number of rules and shorten runtime.

\section{Concluding Remarks}
\label{sec: concluding remarks}
We introduced X-KAN which optimizes multiple local KAN models through an evolutionary framework based on XCSF. By defining local regions via rule antecedents and implementing local KAN models as rule consequents, X-KAN effectively combines KAN's expressiveness with XCSF's adaptive partitioning capability. Our experimental results showed that X-KAN significantly outperforms XCSF, MLP, and KAN for various function approximation problems with 7.2 rules on average. This improvement stems from X-KAN (XCSF)'s principle of assigning fitness based on both accuracy and generality, ensuring high generalization performance.

Future work will explore extending X-KAN as a piecewise symbolic regressor capable of extracting interpretable expressions inspired by \cite{chen2024integrating,liu2024kan}.

\onecolumn
\appendix

\section{Network Architectures of MLPs and KANs}
\label{sec: sup Network Architectures of MLPs and KANs}
\begin{figure*}[h]
    \centering
    \begin{subfigure}[b]{0.49\textwidth}
        \centering
        \includegraphics[width=\textwidth]{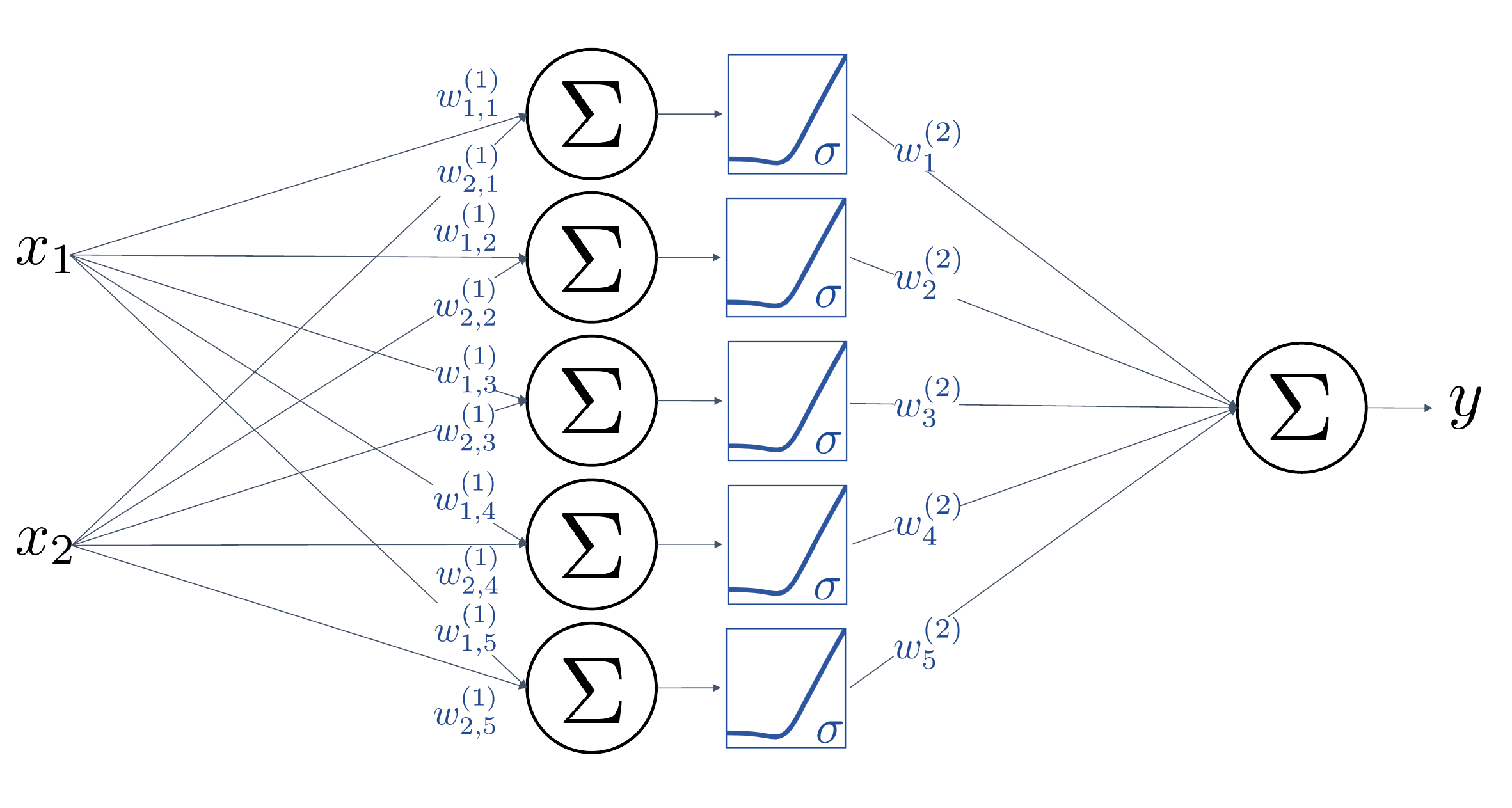}
        \vspace{-3mm}
        \caption{MLPs (cf. Eq. \eqref{eq: mlp} of the main manuscript)}
        \label{fig: sup mlp}
    \end{subfigure}
    \hfill
    \begin{subfigure}[b]{0.49\textwidth}
        \centering
        \includegraphics[width=\textwidth]{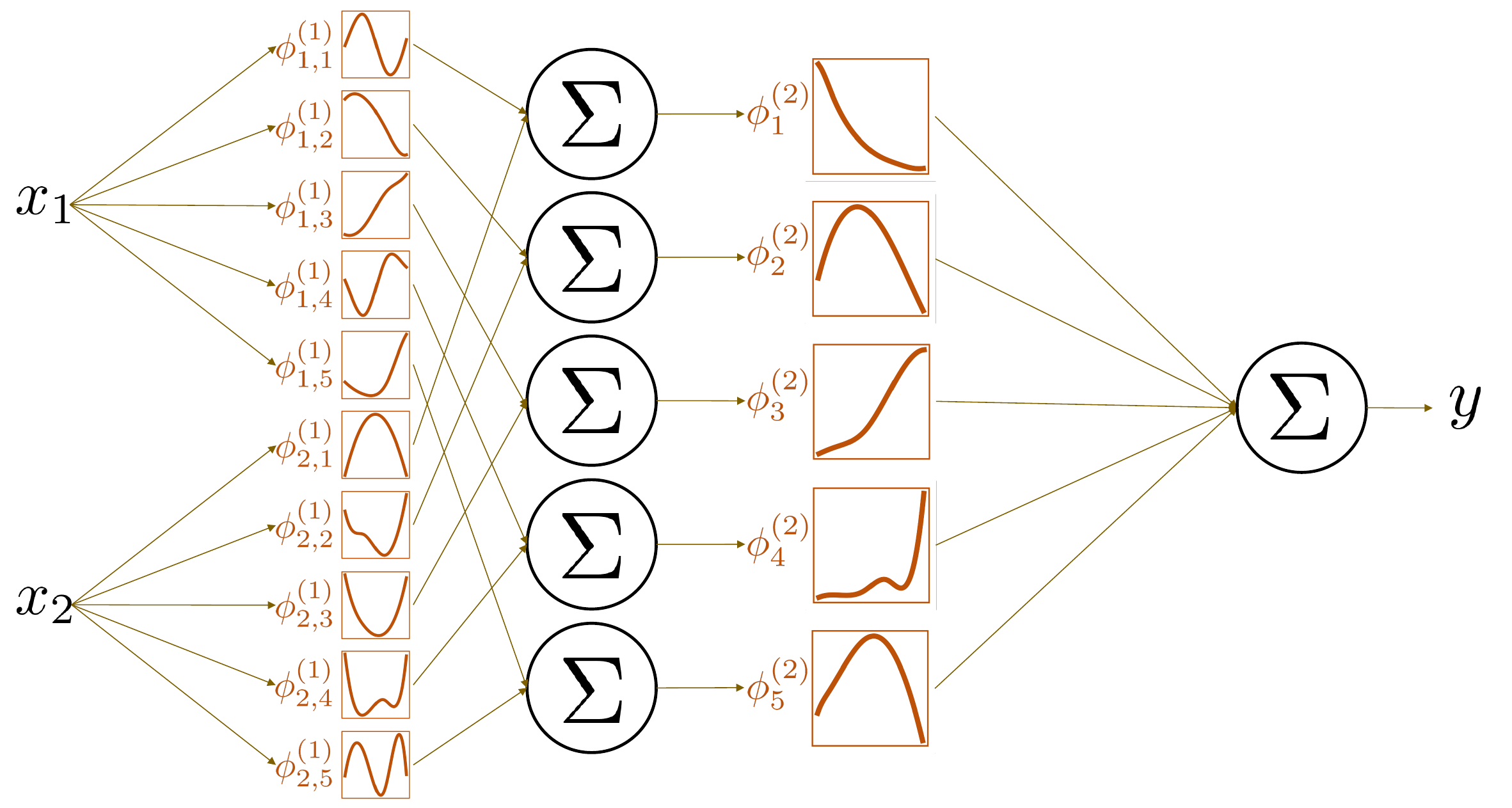}
        \vspace{-3mm}
        \caption{KANs (cf. Eq. \eqref{eq: kan} of the main manuscript)}
        \label{fig: sup kan}
    \end{subfigure}
    \vspace{-2mm}
    \caption{Comparison of MLPs and KANs architectures, both consisting of one input layer, one hidden layer, and one output layer, with input dimension $n=2$. MLPs employ the SiLU function as their activation function. For simplicity of visualization, bias terms in MLPs architectures have been omitted.}
    \label{fig: sup mlp kan}
    \end{figure*}

Fig. \ref{fig: sup mlp kan} illustrates the network architectures of MLPs and KANs. While both MLPs and KANs are fully connected neural networks, their architectures differ significantly.

In MLPs, nodes between layers are connected by learnable weights $w$. Each node generates input features for the next layer by applying a predefined nonlinear activation function (such as Sigmoid \cite{wilson1972excitatory}, ReLU \cite{fukushima1969visual}, or SiLU \cite{elfwing2018sigmoid}) to the sum of the product between the input feature vector and weight matrix plus a bias term. In contrast, KANs employ learnable activation functions $\phi$ on their edges, which combine B-spline functions and SiLU functions, and process information through simple summation of these functions.

Although both models are trained using backpropagation, they differ in their learning targets: MLPs update weight matrices and bias terms, whereas KANs update the control parameters of their activation functions.

\section{Universal Approximation Theorem}
\label{sec: sup Universal Approximation Theorem}
The Universal Approximation Theorem (UAT) \cite{hornik1989multilayer} guarantees the existence of a three-layer MLP that satisfies the following condition for any continuous function $f:\mathbb{R}^n\to\mathbb{R}$ and any tolerance value $\varepsilon\in\mathbb{R}^+$:
\begin{equation}
\sup_{\mathbf{x} \in \mathcal{X} \subseteq \mathbb{R}^n} |f(\mathbf{x}) - \operatorname{MLP}(\mathbf{x})| < \varepsilon.
\end{equation}

UAT theoretically guarantees that a three-layer MLP (cf. Fig. \ref{fig: sup mlp}) with $H>N(\varepsilon)$ nodes in its hidden layer can approximate functions within an error of $\varepsilon$. Note that UAT only demonstrates the possibility of function approximation without specifying the exact value of the required number of nodes $N(\varepsilon)$. Therefore, to improve approximation efficiency, methods such as empirical adjustment of node count, evolutionary optimization of node count (i.e., Neuroevolution \cite{preen2021autoencoding}), or network deepening are essential.

\section{Kolmogorov-Arnold Representation Theorem}
\label{sec: sup Kolmogorov-Arnold Representation Theorem}
The Kolmogorov-Arnold Representation Theorem (KART) \cite{kolmogorov1961representation,arnol1957functions} guarantees that any continuous multivariate function $f$ on a bounded domain can be rewritten as a finite composition of continuous univariate functions and summations. Specifically, $f:[0,1]^n\to \mathbb{R}$ can be expressed as:
\begin{equation}
\label{eq: kart}
    f(\mathbf{x}) =f(x_1,...,x_n)= \sum_{q=1}^{2n+1} \Phi_q\left(\sum_{p=1}^n \phi_{q,p}\left(x_p\right)\right),
\end{equation}
where $\Phi_q: \mathbb{R} \to \mathbb{R}$ and $\phi_{q,p}: [0,1] \to \mathbb{R}$.

Eq. \eqref{eq: kart} indicates that any $n$-variable function can be exactly represented using $2n+1$ univariate functions $\Phi_q$ (outer functions) and $n(2n+1)$ univariate functions $\phi_{q,p}$ (inner functions). Thus, KART enables the simplification of high-dimensional function learning problems into learning multiple, more tractable univariate functions.

\clearpage
\section{XCSF in a Nutshell}
\label{sec: sup Model Architecture of XCSF}

\subsection{Architecture}
\begin{figure}[h]
    \centering
    \includegraphics[width=0.65\linewidth]{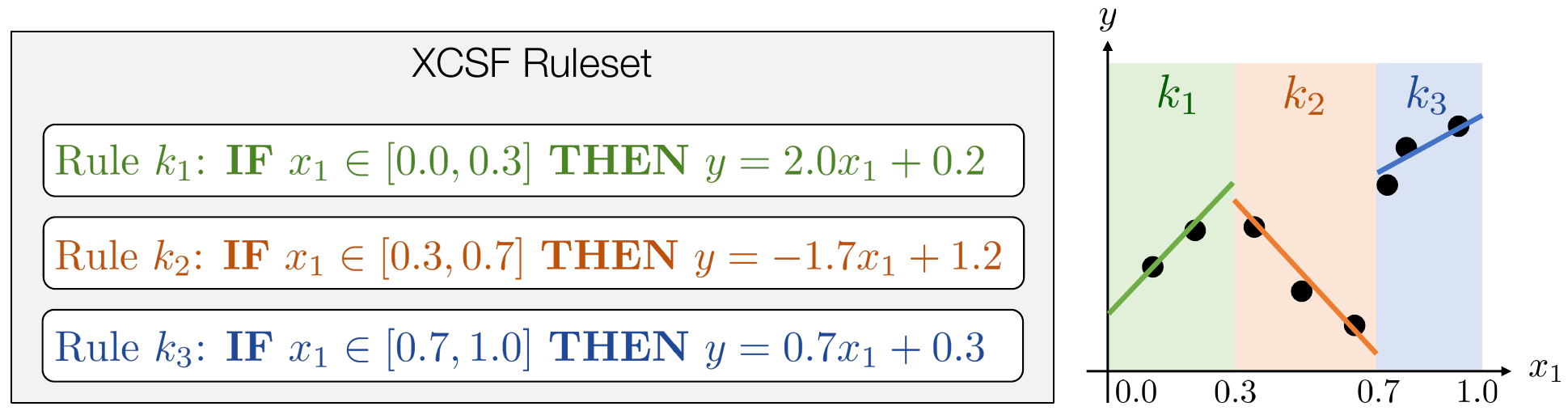}
    \vspace{-2mm}
    \caption{An example of a ruleset of XCSF with three rules, $k_1$, $k_2$, and $k_3$, in an input space $[0,1]$. XCSF partitions the input space into local hyperrectangular regions defined by rule antecedents and performs local function approximation within each region using a linear model implemented in the rule consequent.}
    \label{fig: sup xcsf architecture}
    \end{figure}

Fig. \ref{fig: sup xcsf architecture} schematically illustrates XCSF. XCSF partitions the input space with IF parts (antecedents) and assigns a linear model (consequent) for local approximation. Each rule’s consequent is expressed as:
\begin{equation}
    y=\mathbf{w}\cdot\mathbf{x}+b,
\end{equation}
where $\mathbf{w}$ and $b$ are fitted through gradient descent. XCSF evolves a ruleset of such rules by updating their antecedents and consequents, promoting wide, general coverage and accurate local models.

\subsection{Algorithm}
\subsubsection{Training Phase}
XCSF maintains a population of rules, $\pop$, and sequentially receives data points from a training dataset $\mathcal{D}$ to optimize $\pop$. When a data point $(\mathbf{x}_t, y_t) \in \mathcal{D}$ is received at time $t$, a match set $\ms$ is formed from rules in $\pop$ whose antecedents match $\mathbf{x}_t$. If $\ms$ is empty, a new rule that matches $\mathbf{x}_t$ is generated. Subsequently, the consequents of rules in $\ms$ are updated via gradient descent (see Appendix \ref{ss: batch evaluation protocol} for details), and rule fitness is updated according to the Widrow-Hoff learning rule \cite{widrow1960adaptive}. Additionally, the evolutionary algorithm (EA) is periodically applied to $\ms$. In this process, two parent rules are selected from $\ms$ using tournament selection based on their fitness, and two offspring rules are generated through crossover and mutation. The offspring rules are added to $\pop$ only if they are not subsumed by their parent rules. When the number of rules in $\pop$ exceeds the maximum population size, excess rules are deleted from $\pop$ based on their fitness.

After training is completed, as in X-KAN, the single winner-based rule compaction algorithm \cite{orriols2008fuzzy} is applied to obtain a compact ruleset, $\pc$.

\subsubsection{Testing Phase}
As in X-KAN, when a testing data point $\mathbf{x}_\text{te}$ is given, XCSF selects a single winner rule $k^*$ from a testing match set $\ms_\text{te}=\{\mathbf{x}_\text{te}\in\mathbf{A}^k\mid k\in\pc\}$ and outputs the prediction $\hat{y}$ using the linear model in the consequent of $k^*$.

It should be noted that the original XCSF employs voting-based inference rather than single winner-based inference. Specifically, in voting-based inference, the prediction is obtained as a weighted average of the outputs from all rules in the match set $\ms$. However, to allow a fair comparison with X-KAN and the application of the single winner-based rule compaction, we adopt single winner-based inference in this study.

\subsection{Batch Evaluation Protocol}
\label{ss: batch evaluation protocol}
XCSF, as a Michigan-style LCS, was originally designed for online learning. In this setting, XCSF receives a data point $(\mathbf{x}_t, y_t)$ at each iteration $t$, updates its rules (i.e., linear models) using the new data, and incrementally refines each rule's error via the Widrow-Hoff learning rule throughout the learning process. In contrast, in X-KAN, each rule also receives $(\mathbf{x}_t, y_t)$ at iteration $t$, but rule learning is immediately completed using all available training data points within the rule’s antecedent range, resulting in a uniquely determined rule error. For fair comparison, we adopt the same batch update strategy for the linear models in XCSF, so that each rule's error is also uniquely determined using all available training data points, as in X-KAN.

This batch evaluation protocol eliminates the need for the traditional subsumption condition requiring rules to appear in the match set $\ms$ at least $\theta_{\text{sub}}$ times. In online learning frameworks, this condition ensured sufficient error refinement through incremental Widrow-Hoff updates. However, in this batch learning approach, rule errors are fully determined through complete dataset evaluation, making this frequency-based criterion redundant. Consequently, our implementation removes the $\operatorname{exp}^k \geq \theta_{\text{sub}}$ requirement while maintaining rigorous subsumption checks based on error thresholds and generality criteria.

While this approach resembles Pittsburgh-style LCSs \cite{urbanowicz2017introduction} in terms of batch evaluation, the EA in our framework still operates at the individual rule level. Although stepwise rule evaluation is essential in reinforcement learning settings (e.g., XCS \cite{wilson1995xcs}), immediate batch evaluation of rules with a finite training dataset is well-suited for supervised learning tasks and is employed in state-of-the-art LCS methods such as Survival-LCS \cite{woodward2024survival}.

\clearpage
\section{The Architecture of X-KAN}
\label{sec: sup the architecture of X-KAN} 
\begin{figure}[h]
    \centering
\includegraphics[width=\linewidth]{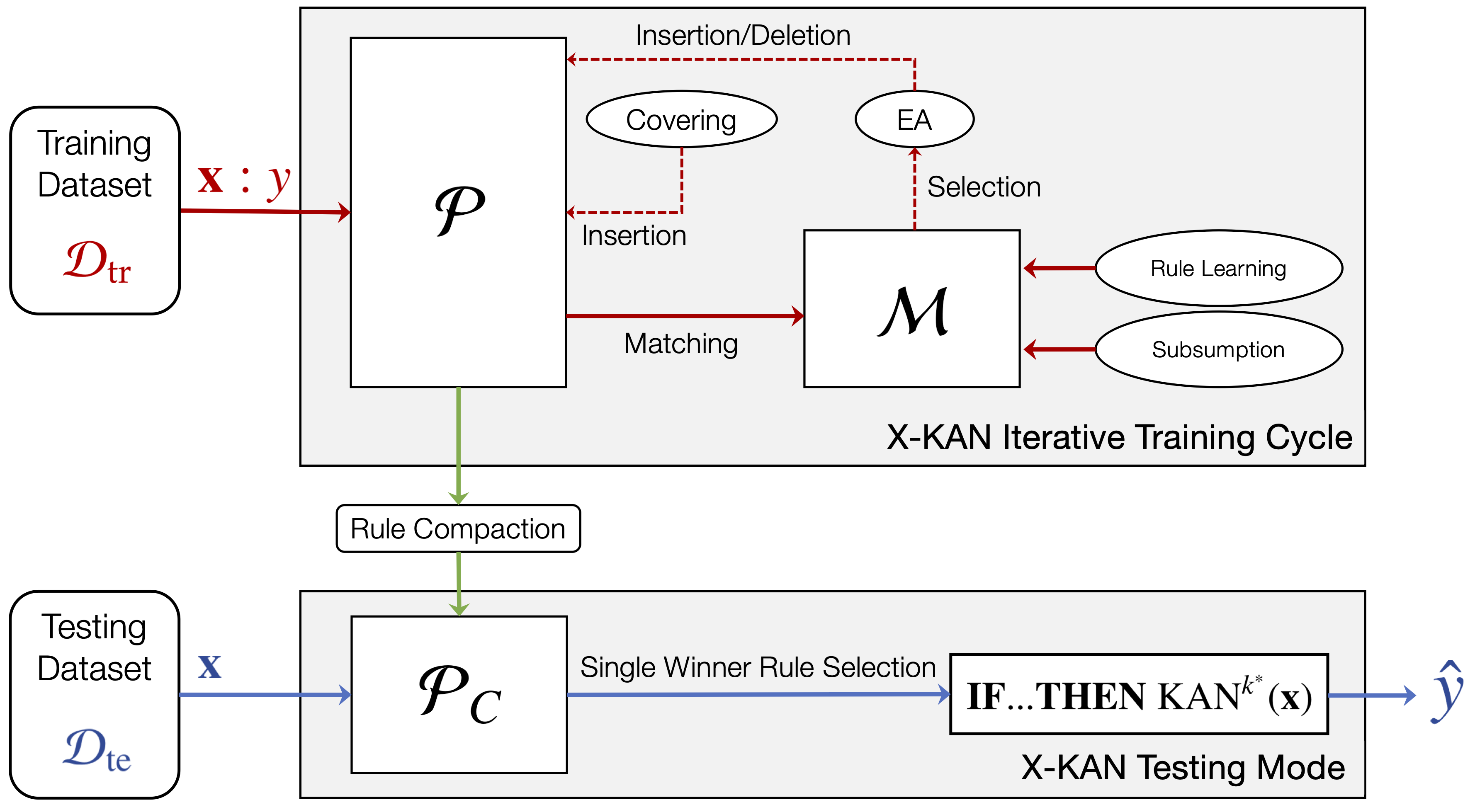}
    \caption{Schematic illustration of X-KAN. The run cycle depends on the type of run: training or testing. Upon receiving each data point $\mathbf{x}$, operations indicated by solid arrows are always performed, while operations indicated by dashed arrows (i.e., covering and GA) are executed only when specific conditions are met.}
    \label{fig: sup x-kan architecture}
    \end{figure}

Fig. \ref{fig: sup x-kan architecture} illustrates the operational workflow of X-KAN through three sequential phases. During the training mode (upper section), the system learns the rule set $\mathcal{P}$ by iteratively processing data points $\mathbf{x}$ from the training dataset $\mathcal{D}_{\text{tr}}$. Subsequently, rule compaction is applied to derive a compact rule set $\mathcal{P}_C$ (middle section), which eliminates redundant rules while preserving approximation accuracy. Finally, in the testing mode (lower section), the compacted rule set $\mathcal{P}_C$ generates predicted values $\hat{y}$ for testing data points $\mathbf{x}$ through single winner-based inference, ensuring efficient and accurate predictions.

\clearpage
\section{Description of the Test Functions}
\label{sec: sup description of the test functions}
The four test functions $f_1$-$f_4$ used in our main manuscript were all employed in \cite{stein2018interpolation}. Each function presents unique difficulties: $f_1$ tests the ability to handle sharp ridges, $f_2$ examines periodic pattern recognition, $f_3$ evaluates performance across mixed complexity regions, and $f_4$ assesses handling of both smooth and steep gradient areas. The remainder of this section details these functions.

\begin{figure*}[b]
    \centering
    \begin{subfigure}[b]{0.24\textwidth}
        \centering
        \includegraphics[width=\textwidth]{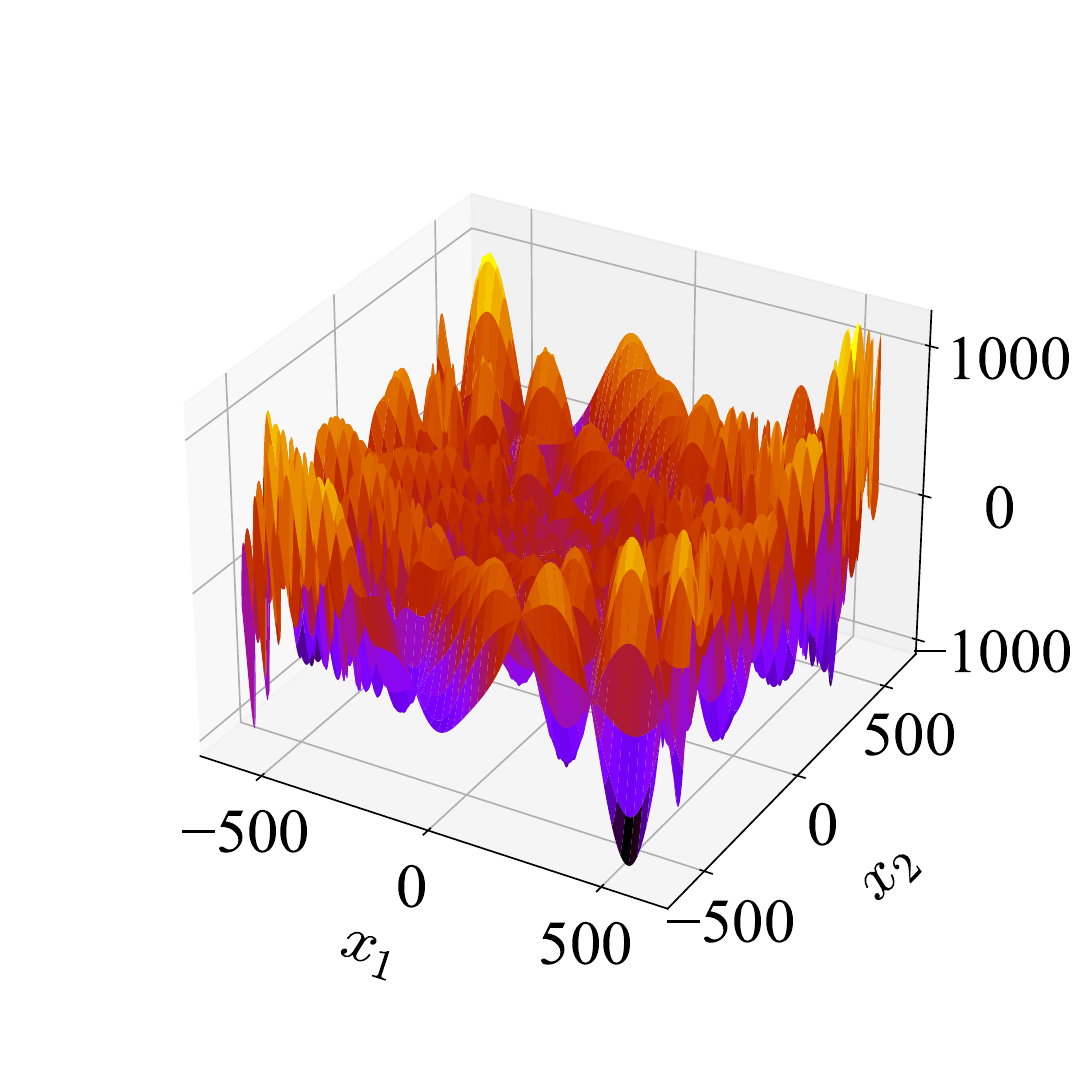}
        \vspace{-3mm}
        \caption{$f_1$: Eggholder function}
        \label{fig: sup egg}
    \end{subfigure}
    \hfill
    \begin{subfigure}[b]{0.24\textwidth}
        \centering
        \includegraphics[width=\textwidth]{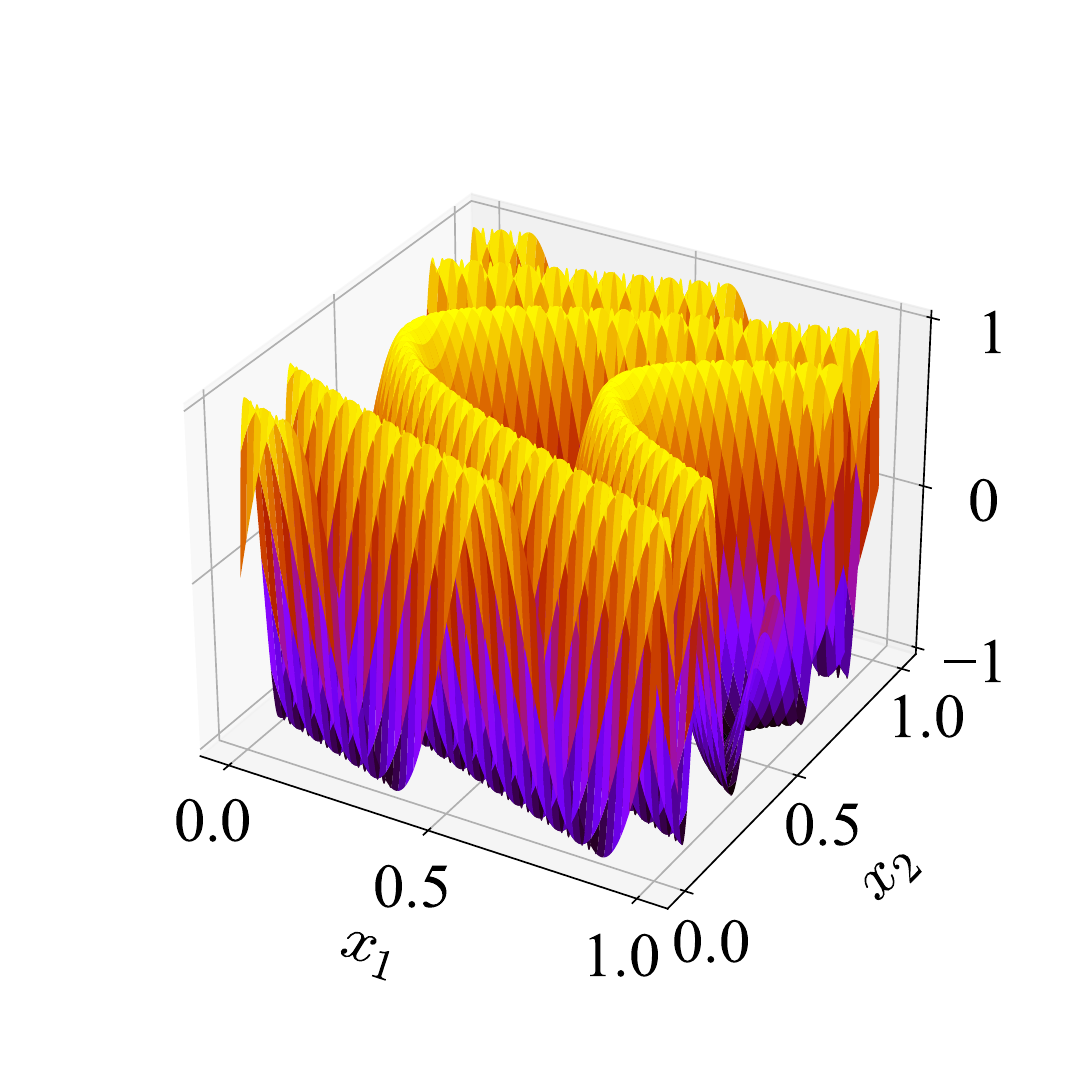}
        \vspace{-3mm}
        \caption{$f_2$: Sine-in-Sine function}
        \label{fig: sup sine}
    \end{subfigure}
    \hfill
    \begin{subfigure}[b]{0.24\textwidth}
        \centering
        \includegraphics[width=\textwidth]{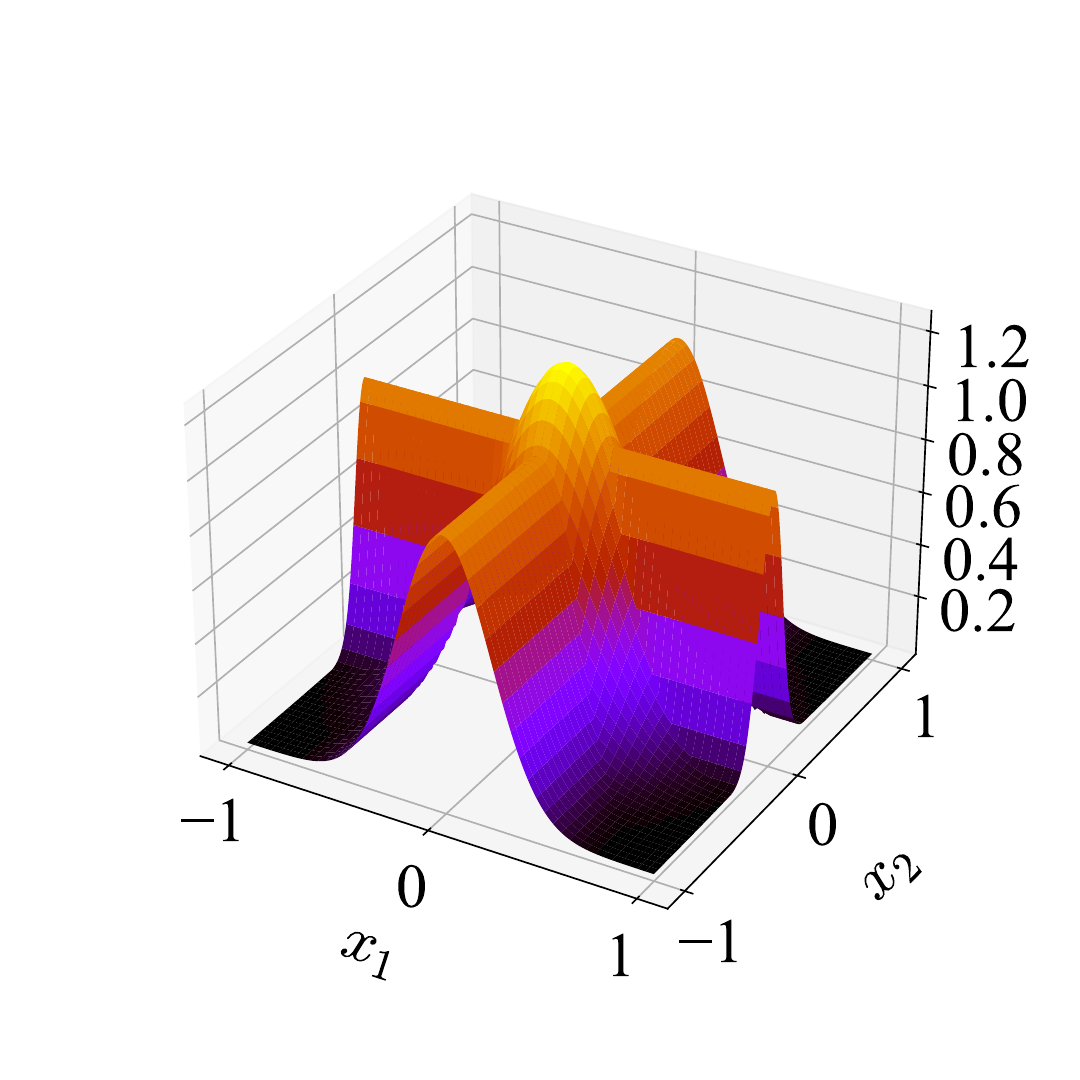}
        \vspace{-3mm}
        \caption{$f_3$: Cross function}
        \label{fig: sup cross}
    \end{subfigure}
    \hfill
    \begin{subfigure}[b]{0.24\textwidth}
        \centering
        \includegraphics[width=\textwidth]{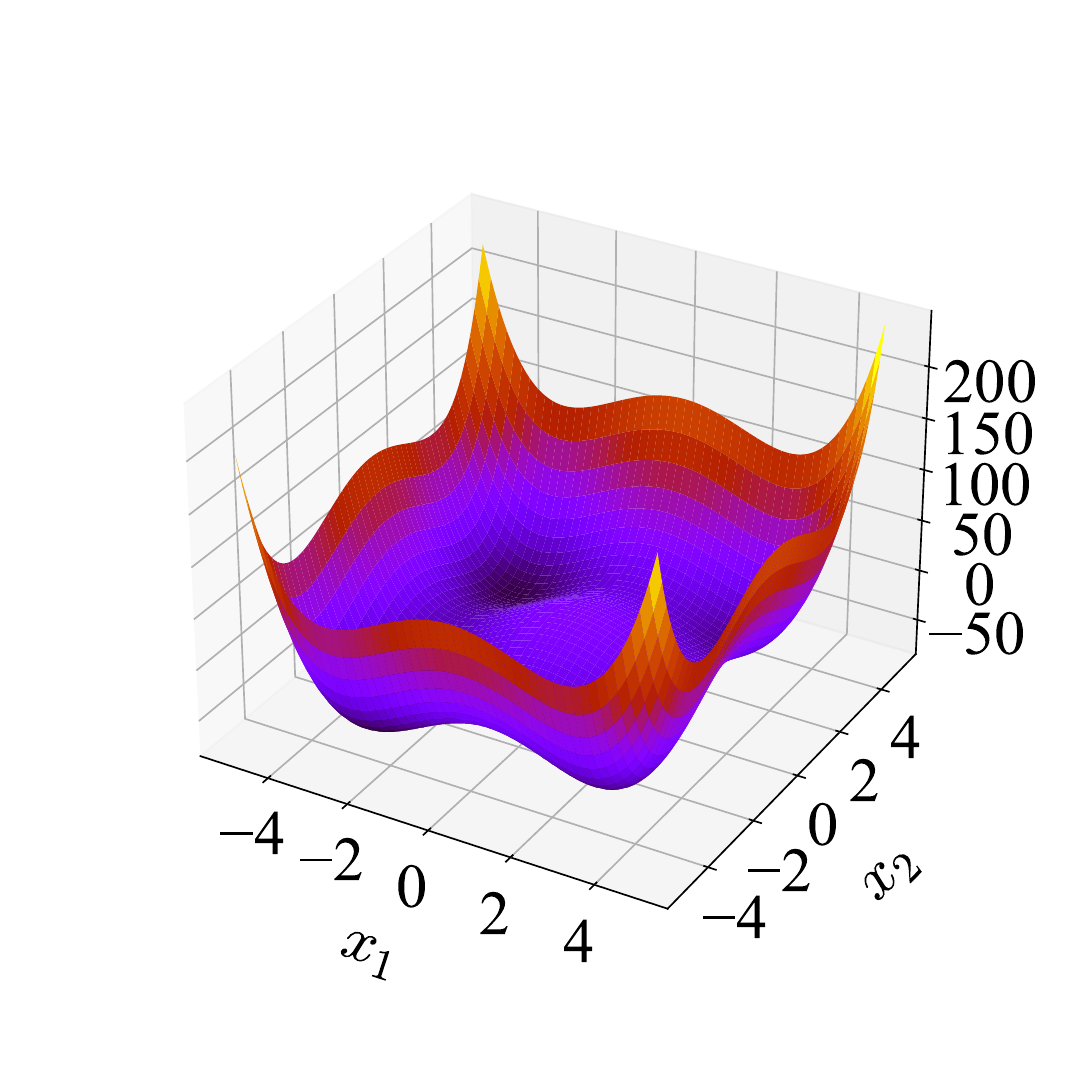}
        \vspace{-3mm}
        \caption{$f_4$: Styblinski-Tang function}
        \label{fig: sup styblinski}
    \end{subfigure}
    \vspace{-2mm}
    \caption{The two-dimensional test functions $f_1$-$f_4$ (same as Fig. \ref{fig: all functions}).}
    \label{fig: sup all functions}
    \end{figure*}

\subsection{$f_1$: Eggholder Function}
The eggholder function is a highly nonlinear function characterized by steep ridges and deep valleys (cf. Fig. \ref{fig: sup egg}):

\begin{align}
f_1(x_1,x_2) = &-(x_2+47)\sin\sqrt{|x_1/2 + (x_2+47)|} \\ \nonumber
&- x_1\sin\sqrt{|x_1-(x_2+47)|},
\end{align}
where:
\begin{equation}
    -512 \leq x_1,x_2 \leq 512.
\end{equation}
This function presents significant challenges for optimization due to its complex surface topology and numerous local optima.

\subsection{$f_2$: Sine-in-Sine Function}
The sine-in-sine function exhibits periodic behavior with interleaved sine waves (cf. Fig. \ref{fig: sup sine}):
\begin{equation}
    f_2(x_1,x_2) = \sin(4\pi(x_1+\sin(\pi x_2))),
\end{equation}
where:
\begin{equation}
    0 \leq x_1,x_2 \leq 1.
\end{equation}
This function features intricate nonlinear patterns and multiple local extrema.

\subsection{$f_3$: Cross Function}
The cross function combines both linear and nonlinear characteristics (cf. Fig. \ref{fig: sup cross}):
\begin{align}
    f_3(x_1,x_2) = \max\{&\exp(-10x_1^2), \exp(-50x_2^2), \\\nonumber
    &1.25\exp(-5(x_1^2+x_2^2))\},
\end{align}
where:
\begin{equation}
    -1 \leq x_1,x_2 \leq 1.
\end{equation}
This function features distinct regions with varying complexity.

\subsection{$f_4$: Styblinski-Tang Function}
The Styblinski-Tang function is a well-known optimization benchmark function (cf. Fig. \ref{fig: sup styblinski}):
\begin{equation}
    f_4(x_1,x_2) = \frac{\sum_{i=1}^{2} x_i^4-16x_i^2+5x_i}{2},
\end{equation}
where:
\begin{equation}
    -5 \leq x_1,x_2 \leq 5.
\end{equation}
This function is characterized by relatively smooth regions in its central domain but exhibits steep gradients near its boundaries.

\clearpage
\section{Description of the Real-World Datasets}
\label{sec: sup description of the real-world}

The four real-world datasets used in our main manuscript were all employed in \cite{heider2023suprb}. These datasets exhibit varying characteristics in terms of linearity, dimensionality, and complexity. The remainder of this section details these datasets.

\subsection{Airfoil Self-Noise (ASN)}
The ASN dataset represents a highly nonlinear regression problem that focuses on the prediction of noise levels generated by airfoils. The dataset contains measurements from a series of aerodynamic and acoustic tests conducted on airfoils in an anechoic wind tunnel.

\subsection{Combined Cycle Power Plant (CCPP) }
The CCPP dataset addresses the prediction of electrical power output from a base load operated combined cycle power plant. The relationship between the features and target variable follows a predominantly linear pattern. The dataset includes measurements of various operational parameters that affect power generation.

\subsection{Concrete Strength (CS)}
The CS dataset deals with the prediction of high-performance concrete compressive strength. It features a complex, nonlinear relationship between its input variables and target value. The dataset incorporates eight input variables that represent different concrete components and age, which influence the final strength characteristics.

\subsection{Energy Efficiency Cooling (EEC)}
The EEC dataset focuses on the prediction of cooling loads in residential buildings. It demonstrates a relatively linear relationship between features and target variables. A notable characteristic of this dataset is its high ratio of input features to samples, which distinguishes it from the other datasets in the collection.

\section{Additional Results}
\label{sec: sup additional results}
\subsection{Effects of Rule Compaction on X-KAN}
\label{ss: sup Effects of Rule Compaction on X-KAN}
Table \ref{tb: rule compaction} compares X-KAN without rule compaction (denoted as ``No Compaction'') and X-KAN with rule compaction (denoted as ``Compaction'') in terms of training MAE, testing MAE, and the number of rules in the ruleset. In other words, X-KAN w/ No Compaction uses $\pop$ for prediction, while X-KAN w/ Compaction uses $\pc$.

The results show identical training MAE values regardless of rule compaction. This consistency is attributable to two factors: the use of single winner rule during prediction, and the compaction process copying all single winner rules from $\pop$ to $\pc$ for every training data point. While testing MAE showed slight degradation for test functions $f_1$ and $f_2$ when applying rule compaction, these differences were not statistically significant. Regarding ruleset size, rule compaction significantly reduced the number of rules across all problems, achieving reductions between 46\% and 72\%.

These findings demonstrate that rule compaction contributes to substantial reduction in total rule count without significantly compromising test approximation accuracy.

\begin{table*}[ht]
    \centering
    \resizebox{\width}{!}{
    \footnotesize
    \begin{tabular}{c |c c|cc|cc}

\bhline{1pt}
& \multicolumn{6}{c}{X-KAN}\\
 & \multicolumn{2}{c|}{\textsc{Training MAE}} & \multicolumn{2}{c|}{\textsc{Testing MAE}} & \multicolumn{2}{c}{\textsc{\#Rules}}\\
  & No Compaction & Compaction & No Compaction & Compaction & No Compaction & Compaction \\
 \bhline{1pt}
$f_1$ & \cellcolor{p}\cellcolor{g}0.10780 $\sim$ & \cellcolor{p}\cellcolor{g}0.10780 & \cellcolor{g}0.16930 $\sim$ & \cellcolor{p}0.16970 & \cellcolor{p}19.03 $-$ & \cellcolor{g}9.200 \\
$f_2$ & \cellcolor{p}\cellcolor{g}0.07356 $\sim$ & \cellcolor{p}\cellcolor{g}0.07356 & \cellcolor{g}0.12710 $\sim$ & \cellcolor{p}0.12730 & \cellcolor{p}19.30 $-$ & \cellcolor{g}10.50 \\
$f_3$ & \cellcolor{p}\cellcolor{g}0.01927 $\sim$ & \cellcolor{p}\cellcolor{g}0.01927 & \cellcolor{p}\cellcolor{g}0.02379 $\sim$ & \cellcolor{p}\cellcolor{g}0.02379 & \cellcolor{p}11.47 $-$ & \cellcolor{g}4.867 \\
$f_4$ & \cellcolor{p}\cellcolor{g}0.05171 $\sim$ & \cellcolor{p}\cellcolor{g}0.05171 & \cellcolor{p}\cellcolor{g}0.06922 $\sim$ & \cellcolor{p}\cellcolor{g}0.06922 & \cellcolor{p}13.97 $-$ & \cellcolor{g}6.400 \\
ASN & \cellcolor{p}\cellcolor{g}0.03548 $\sim$ & \cellcolor{p}\cellcolor{g}0.03548 & \cellcolor{p}\cellcolor{g}0.05533 $\sim$ & \cellcolor{p}\cellcolor{g}0.05533 & \cellcolor{p}24.97 $-$ & \cellcolor{g}7.900 \\
CCPP & \cellcolor{p}\cellcolor{g}0.07427 $\sim$ & \cellcolor{p}\cellcolor{g}0.07427 & \cellcolor{p}\cellcolor{g}0.07871 $\sim$ & \cellcolor{p}\cellcolor{g}0.07871 & \cellcolor{p}22.97 $-$ & \cellcolor{g}7.800 \\
CS & \cellcolor{p}\cellcolor{g}0.02687 $\sim$ & \cellcolor{p}\cellcolor{g}0.02687 & \cellcolor{p}\cellcolor{g}0.07842 $\sim$ & \cellcolor{p}\cellcolor{g}0.07842 & \cellcolor{p}21.93 $-$ & \cellcolor{g}7.900 \\
EEC & \cellcolor{p}\cellcolor{g}0.01561 $\sim$ & \cellcolor{p}\cellcolor{g}0.01561 & \cellcolor{p}\cellcolor{g}0.02729 $\sim$ & \cellcolor{p}\cellcolor{g}0.02729 & \cellcolor{p}9.233 $-$ & \cellcolor{g}2.667 \\
\bhline{1pt}
Rank & \cellcolor{p}\cellcolor{g}\textit{1.50} & \cellcolor{p}\cellcolor{g}\textit{1.50} & \cellcolor{g}\textit{1.38}$\uparrow$ & \cellcolor{p}\textit{1.62} & \cellcolor{p}\textit{2.00}$\downarrow^{\dag}$ & \cellcolor{g}\textit{1.00} \\
Position & \textit{1.5} & \textit{1.5} & \textit{1} & \textit{2} & \textit{2} & \textit{1} \\
Number of $+/-/\sim$ & 0/0/8 & - & 0/0/8 & - & 0/8/0 & - \\
\bhline{1pt}
$p$-value & 1.000 & - & 0.500 & - & 0.00781 & - \\

\bhline{1pt}

\end{tabular}}
    \caption{Summary of experimental results, displaying performance metrics (training and testing MAE and number of rules in the ruleset). Green/peach highlighting indicates best/worst values. Rank and Position indicate the average ranking and final position, respectively. Symbols $+/-/\sim$ denote statistically significant better/worse/similar performance compared to X-KAN with Compaction based on Wilcoxon signed-rank tests. Arrows $\uparrow/\downarrow$ indicate improvement/decline in rank compared to X-KAN with Compaction. Statistical significance at $\alpha=0.05$ is denoted by $\dag$ ($p$-value).}
   
    \label{tb: rule compaction}
\end{table*}

\clearpage
\subsection{Training MAE}
\label{ss: sup training mae}
\begin{figure*}[ht]
    \centering
    \begin{subfigure}[b]{0.24\textwidth}
        \centering
        \includegraphics[width=\textwidth]{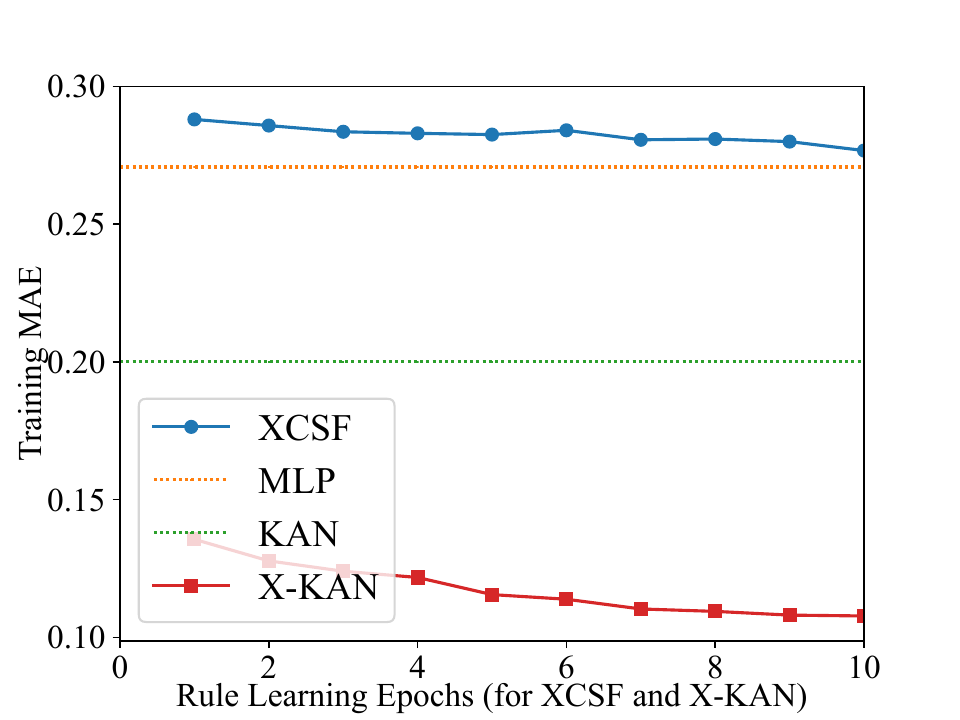}
        \vspace{-3mm}
        \caption{$f_1$: Eggholder function}
        \label{fig: sup f1 train}
    \end{subfigure}
    \hfill
    \begin{subfigure}[b]{0.24\textwidth}
        \centering
        \includegraphics[width=\textwidth]{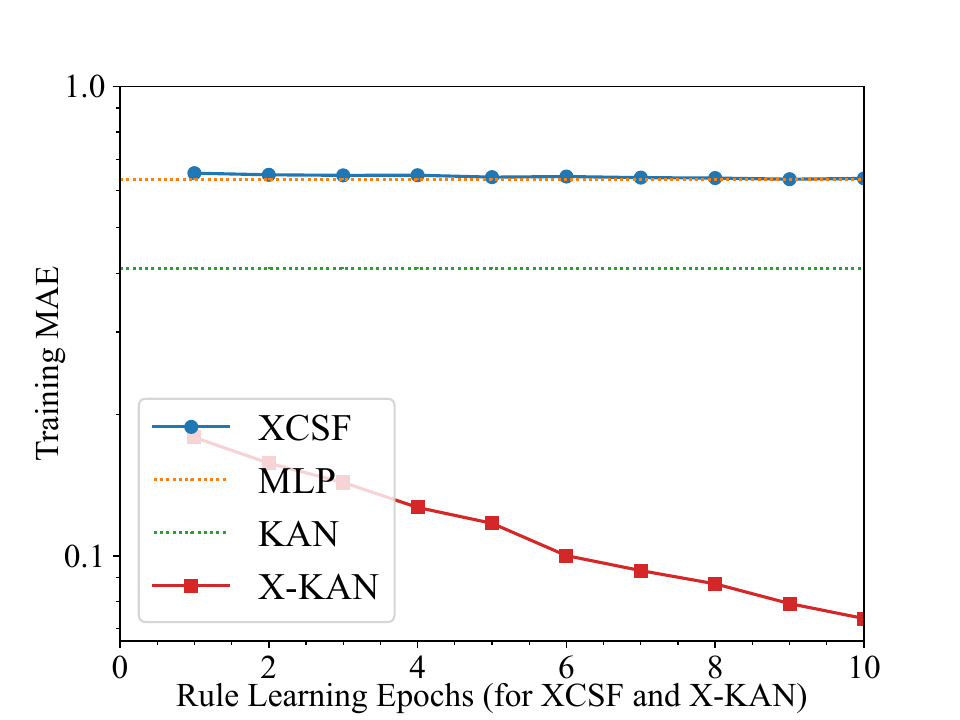}
        \vspace{-3mm}
        \caption{$f_2$: Sine-in-Sine function}
        \label{fig: sup f2 train}
    \end{subfigure}
    \hfill
    \begin{subfigure}[b]{0.24\textwidth}
        \centering
        \includegraphics[width=\textwidth]{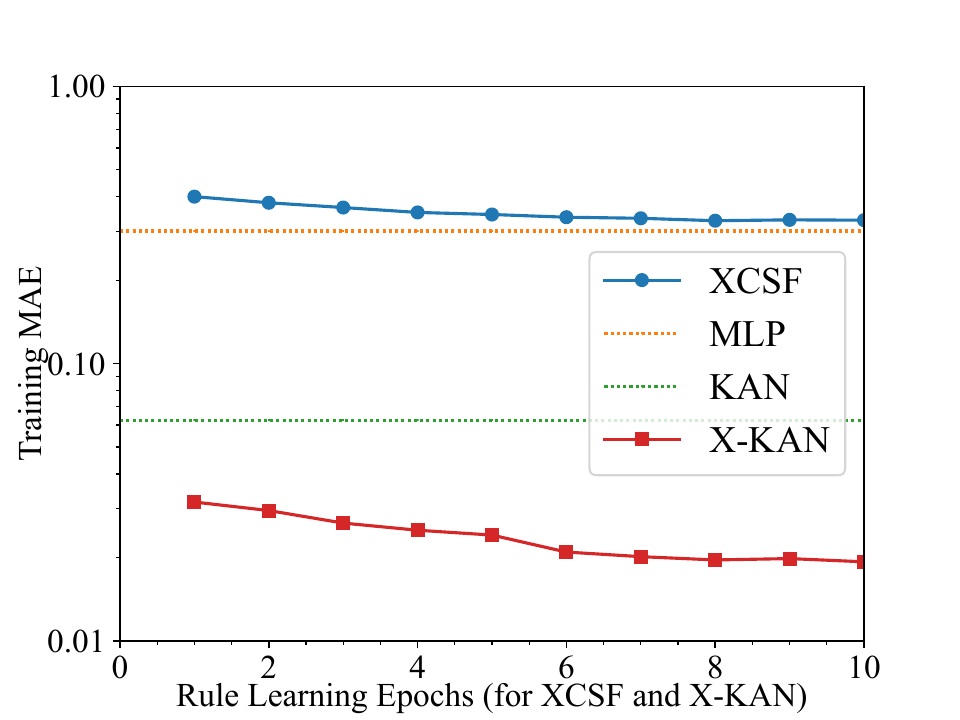}
        \vspace{-3mm}
        \caption{$f_3$: Cross function}
        \label{fig: sup f3 train}
    \end{subfigure}
    \hfill
    \begin{subfigure}[b]{0.24\textwidth}
        \centering
        \includegraphics[width=\textwidth]{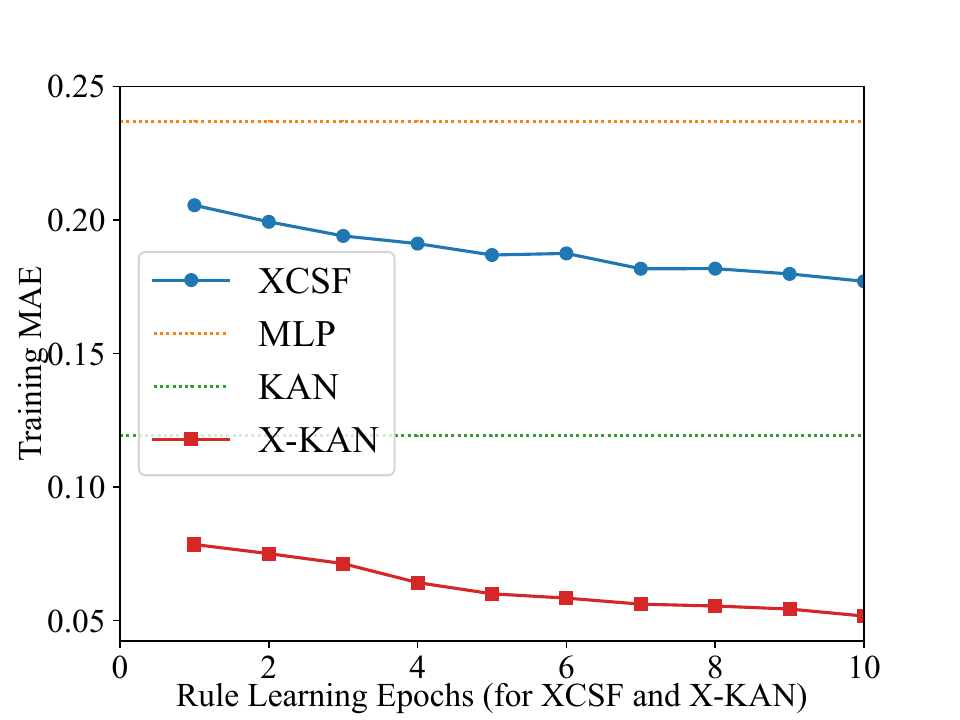}
        \vspace{-3mm}
        \caption{$f_4$: Styblinski-Tang function}
        \label{fig: sup f4 train}
    \end{subfigure}
    \vspace{-2mm}
    \caption{Training MAE curves for the test functions.}
    \label{fig: sup all functions train}
    \end{figure*}

\begin{figure*}[ht]
    \centering
    \begin{subfigure}[b]{0.24\textwidth}
        \centering
        \includegraphics[width=\textwidth]{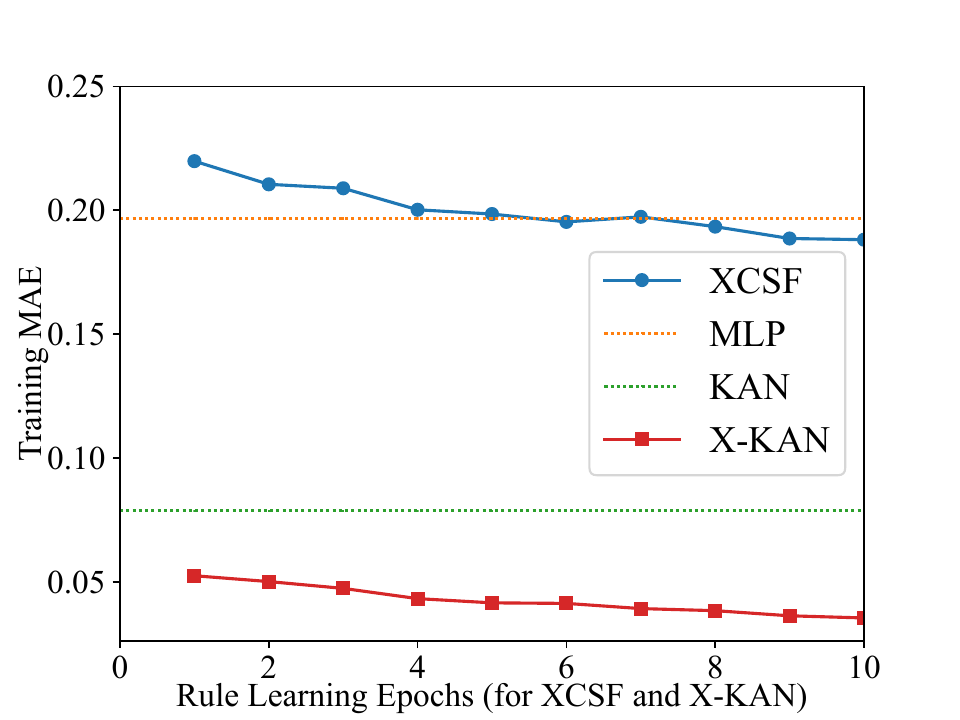}
        \vspace{-3mm}
        \caption{ASN}
        \label{fig: sup asn train}
    \end{subfigure}
    \hfill
    \begin{subfigure}[b]{0.24\textwidth}
        \centering
        \includegraphics[width=\textwidth]{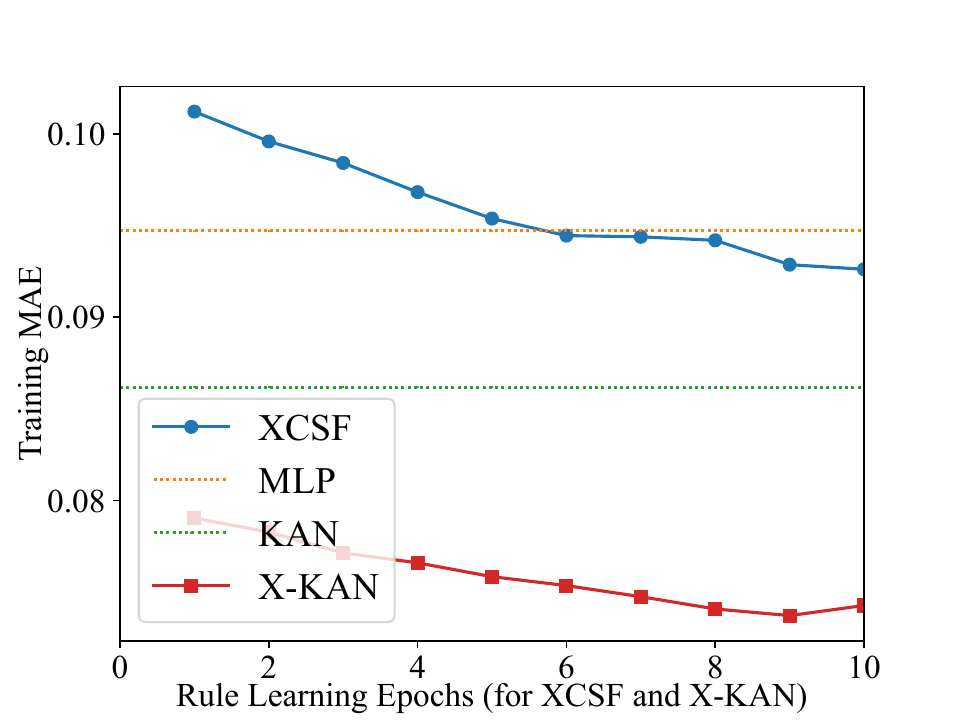}
        \vspace{-3mm}
        \caption{CCPP}
        \label{fig: sup ccpp train}
    \end{subfigure}
    \hfill
    \begin{subfigure}[b]{0.24\textwidth}
        \centering
        \includegraphics[width=\textwidth]{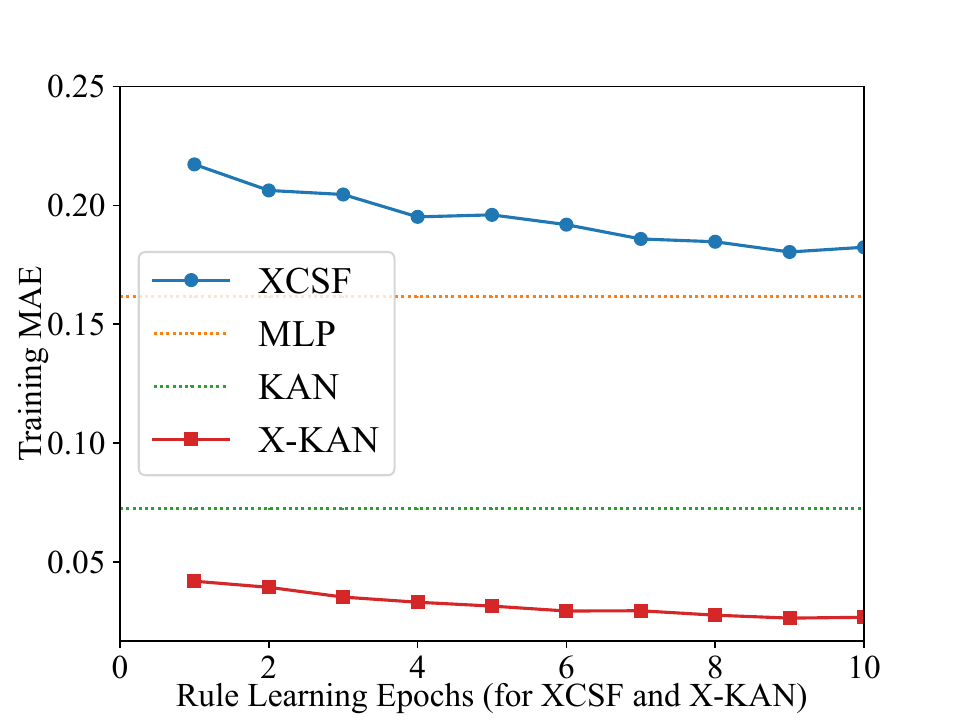}
        \vspace{-3mm}
        \caption{CS}
        \label{fig: sup cs train}
    \end{subfigure}
    \hfill
    \begin{subfigure}[b]{0.24\textwidth}
        \centering
        \includegraphics[width=\textwidth]{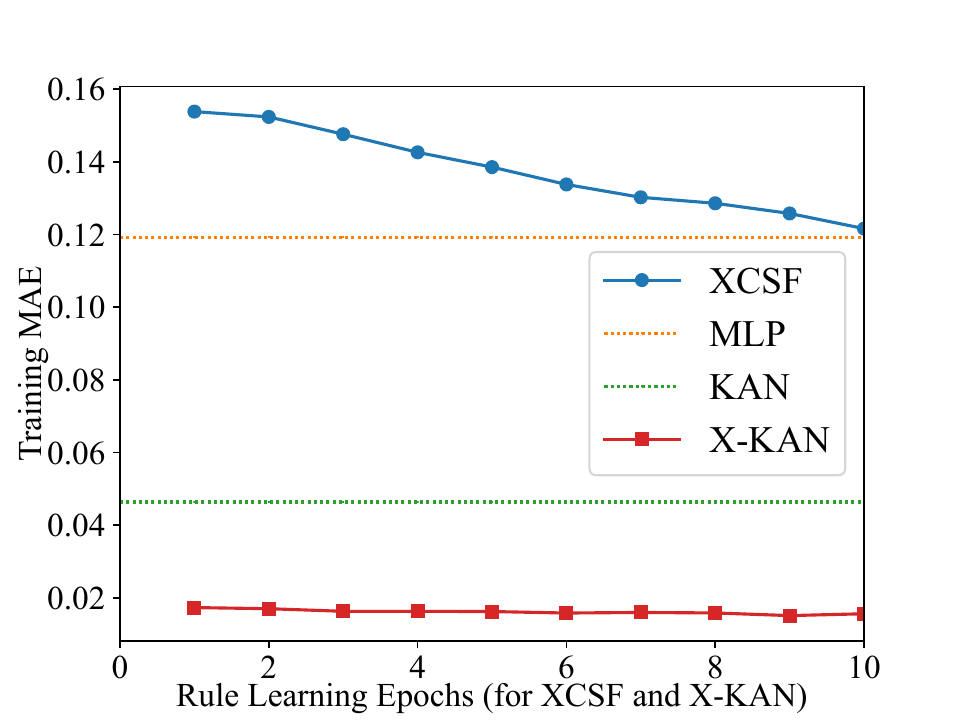}
        \vspace{-3mm}
        \caption{EEC}
        \label{fig: sup eec train}
    \end{subfigure}
    \vspace{-2mm}
    \caption{Training MAE curves for the real-world datasets.}
    \label{fig: sup all real world train}
    \end{figure*}

Figs. \ref{fig: sup all functions train} and \ref{fig: sup all real world train} show the trends of testing MAE during rule learning for XCSF and X-KAN on test functions and real-world datasets, respectively. Final MAE values for MLP and KAN are shown as dashed lines. Both XCSF and X-KAN demonstrate decreasing training MAE as rule learning epochs progress, confirming the effectiveness of local approximation using rules.

\clearpage

\subsection{Comparison with X-MLP}
\label{ss: comparison with x-mlp}
To investigate the generality of our evolutionary rule-based framework, we conducted experiments replacing the local KAN models in X-KAN with MLPs, resulting in X-MLP. In X-MLP, each rule’s consequent is implemented as an MLP with the same architecture and parameter count as the KANs used in X-KAN, ensuring a fair comparison. All other aspects of the framework, including evolutionary optimization of rule antecedents and the fitness assignment mechanism, remain unchanged.

Table \ref{tb: x-mlp and widekan} compares the testing MAE for XCSF, MLP, X-MLP, KAN, WideKAN (described later in the next subsection), and X-KAN. The results demonstrate that X-MLP consistently outperforms the global MLP baseline, achieving lower testing MAE on 7 out of 8 problems. This indicates that the divide-and-conquer approach of our evolutionary rule-based framework is effective regardless of the choice of local approximator, as it enables localized modeling of complex or heterogeneous functions that are difficult for a single global model.

However, X-KAN achieves lower testing MAE than X-MLP on all problems, with statistically significant improvements in most cases. This highlights the particular suitability of KANs as local models within our framework, likely due to their higher parameter efficiency and flexibility in representing nonlinear functions via learnable spline-based activations. These findings are consistent with recent literature that suggests KANs can outperform MLPs in function approximation tasks, especially when model size is constrained \cite{yu2024kan}.

These results confirm that the evolutionary rule-based framework is effective with both MLPs and KANs as local approximators. However, KANs provide a consistent performance and interpretability advantage, supporting their use as the default local model in X-KAN.

\subsection{Comparison with WideKAN}
\label{ss: comparison with widekan}
To ensure a fair comparison in terms of total parameter count, we further evaluated a WideKAN baseline. 
WideKAN with more hidden units than the standard global KAN has the same number of parameters as X-KAN with 50 local KANs (which is the maximum number of local KANs in our experiments).
This comparison addresses the question of whether X-KAN’s superior performance is merely due to having more parameters, or whether the divide-and-conquer strategy and evolutionary partitioning provide additional benefits.

As shown in Table \ref{tb: x-mlp and widekan}, WideKAN achieves better performance than the standard KAN in most cases, reflecting the benefits of increased model capacity. However, X-KAN consistently outperforms WideKAN on all problems, with statistically significant differences in most cases. This indicates that the performance gains of X-KAN cannot be attributed solely to the increased number of parameters. Instead, the evolutionary rule-based search enables X-KAN to partition the input space and assign specialized local models where needed, allowing it to capture local complexities and discontinuities that are difficult for a single, monolithic model to approximate.

These experiments demonstrate that X-KAN’s advantage arises from its evolutionary search and adaptive partitioning, which enable specialized local approximations. This cannot be achieved simply by increasing the size of a single global KAN model. Thus, X-KAN’s design offers a principled and effective approach for function approximation tasks with locally complex or discontinuous structures.

\begin{table}[ht]
    \centering
    \scalebox{1}{
    
    \begin{tabular}{c |cccccc}

\bhline{1pt}
&\multicolumn{6}{c}{\textsc{Testing MAE}}\\
&XCSF&MLP&X-MLP&KAN&WideKAN&X-KAN\\
 \bhline{1pt}
$f_1$ & \cellcolor{p}0.27970 $-$ & 0.27170 $-$ & 0.27180 $-$ & 0.22220 $-$ & 0.20180 $-$ & \cellcolor{g}0.16970 \\
$f_2$ & \cellcolor{p}0.64870 $-$ & 0.63820 $-$ & 0.62070 $-$ & 0.44610 $-$ & 0.27300 $-$ & \cellcolor{g}0.12730 \\
$f_3$ & \cellcolor{p}0.33190 $-$ & 0.30570 $-$ & 0.18580 $-$ & 0.06662 $-$ & 0.04239 $-$ & \cellcolor{g}0.02379 \\
$f_4$ & 0.18070 $-$ & \cellcolor{p}0.24100 $-$ & 0.19110 $-$ & 0.12850 $-$ & 0.11500 $-$ & \cellcolor{g}0.06922 \\
ASN & 0.19240 $-$ & \cellcolor{p}0.19990 $-$ & 0.17750 $-$ & 0.08407 $-$ & 0.06553 $-$ & \cellcolor{g}0.05533 \\
CCPP & 0.09288 $-$ & \cellcolor{p}0.09521 $-$ & 0.09258 $-$ & 0.08684 $-$ & 0.07999 $\sim$ & \cellcolor{g}0.07871 \\
CS & \cellcolor{p}0.19050 $-$ & 0.16490 $-$ & 0.15310 $-$ & 0.08833 $-$ & \cellcolor{g}0.07653 $\sim$ & 0.07842 \\
EEC & \cellcolor{p}0.12560 $-$ & 0.12290 $-$ & 0.11230 $-$ & 0.05232 $-$ & 0.03728 $-$ & \cellcolor{g}0.02729 \\
\bhline{1pt}
Rank & \cellcolor{p}\textit{5.50}$\downarrow^{\dag\dag}$ & \textit{5.25}$\downarrow^{\dag\dag}$ & \textit{4.25}$\downarrow^{\dag\dag}$ & \textit{3.00}$\downarrow^{\dag\dag}$ & \textit{1.88}$\downarrow^{\dag\dag}$ & \cellcolor{g}\textit{1.12} \\
Position & \textit{6} & \textit{5} & \textit{4} & \textit{3} & \textit{2} & \textit{1} \\
$+/-/\sim$ & 0/8/0 & 0/8/0 & 0/8/0 & 0/8/0 & 0/6/2 & - \\
\bhline{1pt}
$p$-value & 0.00781 & 0.00781 & 0.00781 & 0.00781 & 0.0234 & - \\
$p_\text{Holm}$-value & 0.0391 & 0.0391 & 0.0391 & 0.0391 & 0.0391 & - \\
\bhline{1pt}
    
\end{tabular}}
    \caption{Results of testing MAE. Notations follow Table \ref{tb: rule compaction}.}
    \label{tb: x-mlp and widekan}
\end{table}

\clearpage
\section{Description of the Discontinuous Function}
\label{sec: sup description of the discontinuous function}
\begin{figure}[ht]
\centerline{\includegraphics[width=0.5\linewidth]{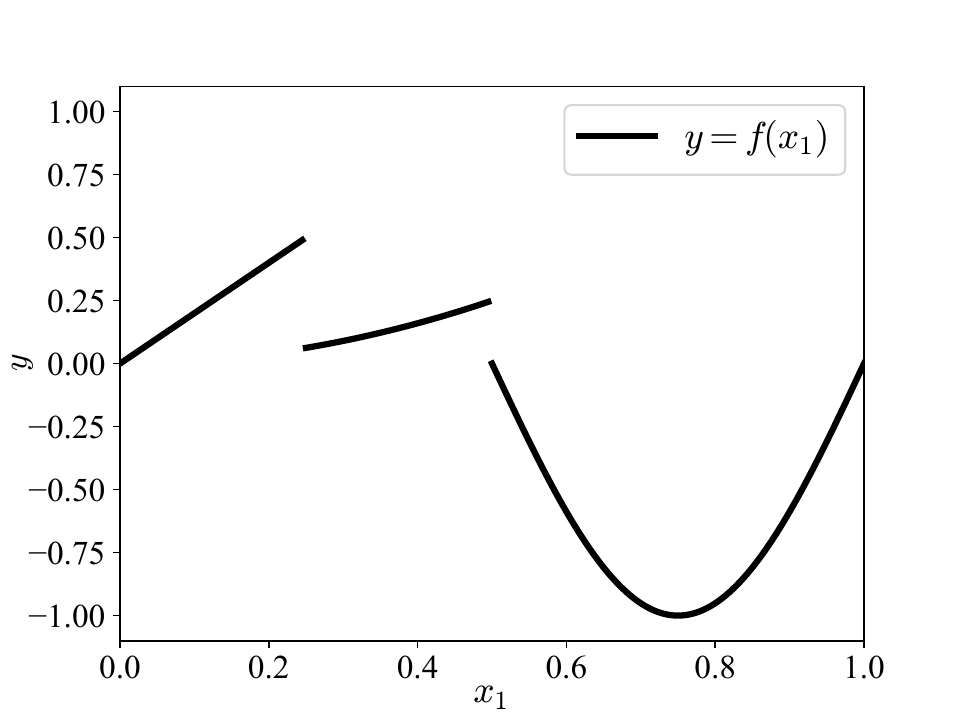}}
\caption{The discontinuous function $f$ used in Section \ref{ss: Analysis on a Discontinuous Function}}
\label{fig: sup piecewise}
\end{figure}

The discontinuous function $f$ used in the main manuscript was employed in \cite{shoji2023piecewise}. It is defined as (cf. Fig. \ref{fig: sup piecewise}):
\begin{equation}
\label{eq: sup discontinuous function}
    f(x_1)=\begin{cases}
2x_1 & \text{if}\ x_1\in[0,0.25),\\
x_1^2 & \text{if}\ x_1\in[0.25,0.5),\\
\sin(2\pi x_1) & \text{if}\ x_1\in[0.5,1],
\end{cases}
\end{equation}
where:
\begin{equation}
    0\leq x_1\leq 1.
\end{equation}

Note that, as described in Section \ref{ss: experimental setup} of the main manuscript, we uniformly sampled 1,000 points from this function and normalized the data targets to the range $[-1,1]$ for the analysis in Section \ref{ss: Analysis on a Discontinuous Function} (cf. Fig. \ref{fig: ground truth}).

\twocolumn
\section*{Acknowledgments}
This work was supported by JSPS KAKENHI (Grant Nos. JP23KJ0993, JP25K03195), National Natural Science Foundation of China (Grant No. 62376115), and Guangdong Provincial Key Laboratory (Grant No. 2020B121201001).
\bibliographystyle{named}
\bibliography{arXiv}
\end{document}